\documentclass[10pt,final, twocolumn]{IEEEtran}
\IEEEoverridecommandlockouts

\usepackage{cite}
\usepackage{amsmath,amssymb,amsfonts}
\usepackage{algorithmic}
\usepackage{graphicx}
\usepackage{textcomp}
\usepackage{xcolor}
\def\BibTeX{{\rm B\kern-.05em{\sc i\kern-.025em b}\kern-.08em
    T\kern-.1667em\lower.7ex\hbox{E}\kern-.125emX}}

\usepackage{acronym}
\usepackage[belowskip=-10pt,aboveskip=1.8pt]{caption}
\usepackage{standalone}
\usepackage{subcaption}
\usepackage{tikz}
\usepackage{url,enumitem, cite}
\usepackage{verbatim}
\usepackage[bookmarks,colorlinks]{hyperref}
\usepackage{soul, xcolor}
\usepackage{mathrsfs}
\usepackage[ruled,linesnumbered,vlined]{algorithm2e}
\SetKwInput{KwData}{\textbf{Init}}

\usepackage[all=normal,paragraphs=tight,floats=normal,mathspacing=normal,wordspacing=tight,charwidths=tight,mathdisplays=normal,leading=normal]{savetrees}

\newcommand{\myVec}[1]{{\boldsymbol{#1}}}
\newcommand{\myMat}[1]{{\boldsymbol{#1}}}
\newcommand{\mySet}[1]{\mathcal{#1}} 
\newcommand{\delt}[2]{\myVec{\delta}_{#1}^{#2}}
\newcommand{\normtwo}[1]{\left\|#1\right\|_2}

\newtheorem{definition}{Definition}

\newtheorem{proposition}{Proposition}

\let\oldnl\nl
\newcommand{\nonl}{\renewcommand{\nl}{\let\nl\oldnl}}

\acrodef{cw}[CW]{Carlini Wagner}
\acrodef{mse}[MSE]{mean-squared error}
\acrodef{mimo}[MIMO]{multiple-input multiple-output}
\acrodef{cs}[CS]{compressed sensing}
\acrodef{ml}[ML]{machine learning}
\acrodef{dnn}[DNN]{deep neural network}
\acrodef{rmse}[RMSE]{root mean squared error}
\acrodef{rmspe}[RMSPE]{root mean squared periodic error}
\acrodef{mmse}[MMSE]{{minimum mean-squared error}}
\acrodef{lmmse}[LMMSE]{{linear} MMSE}
\acrodef{mle}[MLE]{maximum likelihood estimation}
\acrodef{snr}[SNR]{signal-to-noise ratio}
\acrodef{gd}[GD]{gradient descent}
\acrodef{ista}[ISTA]{iterative soft thresholding algorithm}
\acrodef{lista}[LISTA]{learned iterative soft thresholding algorithm}
\acrodef{pca}[PCA]{principal component analysis}
\acrodef{svd}[SVD]{singular value decomposition}
\acrodef{gan}[GAN]{generative adversarial network}
\acrodef{gcn}[GCN]{graph convolutional network}
\acrodef{admm}[ADMM]{alternating direction method of multipliers}
\acrodef{rpca}[RPCA]{robust principal component analysis}  
\acrodef{fgsm}[FGSM]{fast gradient sign method}
\acrodef{nifgsm}[NIFGSM]{Nesterov iterative \ac{fgsm}}
\acrodef{pgd}[PGD]{projected gradient descent}

\acrodef{bim}[BIM]{basic iterative method}
\acrodef{lasso}[LASSO]{least absolute shrinkage and selection operator}

\setstcolor{blue}

\begin{document}

\title{Unveiling and Mitigating Adversarial Vulnerabilities\\ in Iterative Optimizers}

\author{Elad Sofer, Tomer Shaked, Caroline Chaux, and Nir Shlezinger
\thanks{Parts of this work were presented at the  International Workshop on
Machine Learning for Signal Processing (MLSP) 2023 as the paper \cite{sofer2023interpretable}.
E. Sofer, T. Shaked, and N. Shlezinger are with the School of ECE, Ben-Gurion University of the Negev, Be’er-Sheva, Israel (e-mails: \{eladsofe; tosha\}@post.bgu.ac.il;
nirshl@bgu.ac.il). C. Chaux is with CNRS, IPAL, Singapore (e-mail: caroline.chaux@cnrs.fr).}}

\maketitle
\begin{abstract}

\Ac{ml} models are often sensitive to carefully crafted yet seemingly unnoticeable perturbations. Such adversarial examples are considered to be a property of \ac{ml} models, often associated with their black-box operation and sensitivity to features learned from data. 
This work examines the adversarial sensitivity of non-learned decision rules, and particularly  of iterative optimizers. Our analysis is inspired by the recent developments in deep unfolding, which cast such optimizers as \ac{ml} models. 
We show that non-learned iterative optimizers share the sensitivity to adversarial examples of \ac{ml} models, and that attacking iterative optimizers effectively alters the optimization objective surface in a manner that modifies the minima sought. We then leverage the ability to cast iteration-limited optimizers as \ac{ml} models  to enhance robustness via adversarial training. For a class of proximal gradient optimizers, we rigorously prove how their learning affects adversarial sensitivity.  We numerically back our findings, showing the vulnerability of various  optimizers, as well as the robustness induced by unfolding and adversarial training.   
\end{abstract}

\acresetall
 

\section{Introduction}
\label{sec:intro}
Adversarial examples constitute a growing threat to \ac{ml} models~\cite{szegedy2013intriguing}, and are the focus of an ongoing arms race involving the development of sophisticated adversarial attacks along with  countermeasures that mitigate  sensitivity~\cite{akhtar2021advances, wang2023adversarial}.  Vulnerability to such attacks, which are based on crafting minor, seemingly unnoticeable perturbations added to an input signal, is typically viewed as a property of \ac{ml}-based inference rules, with the most common type of vulnerable models being \acp{dnn}~\cite{silva2020opportunities}. While there is no clear consensus on why adversarial examples exist~\cite{ignatiev2019relating,zhang2020understanding,li2024adversarial}, vulnerability is often attributed to the  non-interpretable operation  of \acp{dnn}, and their sensitivity to certain features learned in training~\cite{ilyas2019adversarial}.

  An alternative approach to \ac{ml} is to design inference mappings is based on principled mathematical descriptions of the task. One such family of inference rules, that is well trusted and widely used in various domains~\cite{luo2006introduction}, is based on formulating  a closed-form optimization  problem and tackling it using iterative solvers~\cite{boyd2004convex}. 
Iterative  optimizers fundamentally differ from \ac{ml} models, as they are not learned from data, and do not share the notion of generalizations. However, there exist certain similarities between  iterative optimizers and \acp{dnn}~\cite{shlezinger2022model}. These similarities are  exploited in methodologies that convert  optimizers into \ac{ml} models~\cite{shlezinger2023model}. 

In particular, the casting of iterative optimizers as \ac{ml} models gained attention with the introduction of \ac{lista} in \cite{gregor2010learning}. \ac{lista} and its numerous extensions~\cite{sprechmann2015learning,xiang2021fista,chen2021hyperparameter,aberdam2021ada} demonstrated that iterative optimizers can be reformulated as \ac{ml} models by fixing a number of iterations and learning their hyperparameters, treating each iteration as a layer of a \ac{dnn}. This concept, termed deep unfolding~\cite{monga2021algorithm}, has led to extensive research into leveraging differentiability and structured priors inherent in optimization methods to improve  performance through data-driven training~\cite{hershey2014deep, diamond2017unrolled, agrawal2021learning, shlezinger2022model,chen2022learning, shlezinger2023model}. However, the ability to treat iterative optimizers as \ac{ml} models indicates that they may share the sensitivity to adversarial attacks of \ac{ml} models.

Traditionally, robustness in iterative optimization is analyzed with respect to noise or outliers~\cite{beyer2007robust}, which are  independent of the input. This focus stems from the use of iterative optimizers in tasks where data are  noisy. While some popular  optimizers such as \ac{admm} were shown to have some level of robustness to noise~\cite{boyd2011distributed, li2022robust}, one can enhance robustness by altering the objective function, e.g., employing outlier-insensitive measures (e.g., the Huber loss~\cite{huber1964robust}), or via regularization~\cite{tibshirani1996regression}, see also survey in \cite{gabrel2014recent}.  However, these methods typically do not account for input-specific tailored perturbations, such as adversarial attacks. Tackling input-specific perturbations typically necessitates modifying the overall optimization framework, e.g., replacing a minimization problem with a minimax setup~\cite{beyer2007robust}, which in turn requires dedicated iterative methods that differ from ones designed for standard minimization~\cite{gorissen2015practical, thekumparampil2019efficient}.

\begin{figure} 
    \centering
    \includegraphics[width=\linewidth]{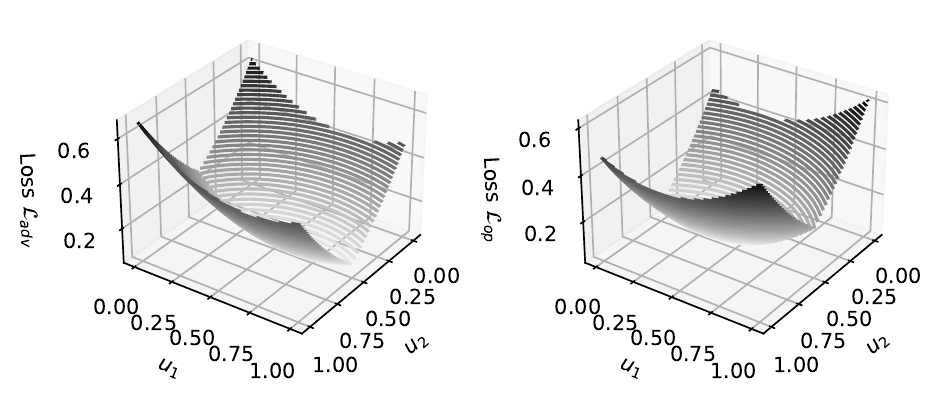}
    \caption{2D projected loss surfaces  of attacked proximal gradient descent (left) and  clean proximal gradient descent (right)} 
    \label{fig:ISTALoss}
\end{figure}

The study of iterative optimizers in the context of adversarial attacks has largely focused on their role in training \ac{ml} models, rather than their application to inference. These include studies that investigate the interplay between optimizers employed in training and robustness to adversarial perturbations of the trained \ac{dnn}~\cite{wang2019assessing, zhang2020understanding}. Studies including \cite{madry2017towards,lee2021towards, zhang2022revisiting} demonstrated that adversarial training with appropriately tuned iterative optimizers significantly enhances robustness, establishing it as a benchmark for robust training methodologies.  Concurrently, \cite{liang2023optimization} explored how iterative optimizers could be tailored to design robust training schemes that mitigate adversarial vulnerabilities by incorporating regularization or adaptive learning rates. Despite this progress, there is limited exploration of adversarial sensitivity in iterative optimizers used directly as inference rules. This gap motivates a shift in perspective, addressing adversarial attacks on iterative methods employed for solving optimization problems rather than for training \ac{ml} models.

In this work we explore two core research questions arising from the ability to cast iterative optimizers as \acp{dnn}: 
$(i)$ in what sense do iterative optimizers share the vulnerability of such \ac{ml} models to adversarial examples? 
and $(ii)$ how does the learning of unfolded optimizers affect sensitivity to adversarial examples?  
We focus on inference rules implemented via descent methods designed for convex optimization, and show that the effect of adversarial attacks, which in the traditional context of \acp{dnn} is somewhat elusive, can be translated into {\em altering the optimization objective} when attacking iterative optimizers, as illustrated in Fig.~\ref{fig:ISTALoss}.  This behavior is experimentally demonstrated for several  optimizers for \ac{cs}, \ac{rpca},  and hybrid beamforming. 

To understand the effect of unfolding on adversarial sensitivity, we adopt the established relationship between Lipschitz continuity and adversarial sensitivity~\cite{zuhlke2024adversarial}. For the representative families of  optimizers based on both proximal \ac{gd} and \ac{admm} with a linear data matching term, we rigorously relate the learned parameters of unfolded algorithms with the Lipschitz constant. This characterization reveals how learning affects vulnerability in unfolded optimizers, while indicating on the ability to exploit their casting as \ac{ml} models to enhance robustness via adversarial training. Our numerical study systematically shows that sensitivity is alleviated via adversarial training. 

%

 The rest of this work is organized as follows:
 Section~\ref{sec:Adversarial} reviews necessary preliminaries in adversarial examples. In Section~\ref{sec:Attacking}, we analyze the vulnerability of iterative optimizers to adversarial attacks, along with the ability to robustify them via adversarial-aware deep unfolding. Our experimental study is presented in Section~\ref{sec:sims}, while Section~\ref{sec:conclusions} provides concluding remarks.

%
\section{Adversarial Examples}\label{sec:Adversarial} 
\subsection{Formulation}\label{ssec:AdversarialForm} 
Adversarial examples are carefully crafted perturbations made to an input in order to affect the decision rule of an \ac{ml} model. Such seemingly imperceptible perturbations can cause \ac{ml} models, e.g., a trained \ac{dnn}, to notably deteriorate their performance~\cite{silva2020opportunities}. 
To formulate their operation, consider an \ac{ml} model with parameters $\myVec{\theta}$ denoted $f_{\myVec{\theta}}(\cdot)$ that maps an observation $\myVec{x}$ from an input space $\mySet{X}$ into an estimate $\myVec{s}$ taking values in a target space $\mySet{S}$.  The task is dictated by a loss function $\mySet{L}$, e.g., a supervised empirical risk  over the training data. An adversarial example is a perturbation $\myVec{\delta}$ taking values in a set of allowed perturbations $\Delta \subset \mySet{X}$,  designed to {\em maximize} the loss for  data $\mySet{D}_{e}$, namely, 
\begin{equation}
    \myVec{\delta}^{\star} = \mathop{\arg \max}\limits_{\myVec{\delta} \in \Delta} \frac{1}{|\mySet{D}_e|} \sum_{(\myVec{x},\myVec{s})\in \mySet{D}_e}\mySet{L}(f_{\myVec{\theta}}, \myVec{x} + \myVec{\delta}, \myVec{s}). 
    \label{eqn:AdvGoal}
\end{equation}

The set $\Delta$  contains only minor perturbations, typically  vectors whose $\ell_p$ norm is bounded by some small $\epsilon>0$ (which we denote here by $\Delta_p(\epsilon)$). The set $\mySet{D}_e$  often includes a single $(\myVec{x},\myVec{s})$ pair, as  considered in the sequel, though adversarial examples can be designed for  $|\mySet{D}_e|>1$. 

\subsection{Attack Algorithms}\label{ssec:AdversarialAlg}
Various  algorithms have been proposed to design adversarial perturbations $\myVec{\delta}$ based on \eqref{eqn:AdvGoal} (detailed examples for different tasks and loss measures $\mySet{L}(\cdot)$ are given in Section~\ref{sec:sims}). We next present two representative basic methods that are relevant in the context of  iterative optimizers, focusing on attacks that have access to the model $f_{\myVec{\theta}}$, i.e., white-box attacks.  

{\em \Ac{fgsm}} introduced in \cite{goodfellow2014explaining} is a basic scheme for generating adversarial examples. \ac{fgsm} generates perturbations by advancing each coordinate  based on the gradient sign of \eqref{eqn:AdvGoal}, i.e., 
\begin{equation}
    \label{eqn:fgsp}
     \myVec{\delta} =  \epsilon  \cdot \text{sign}\Big(
     \nabla_{\myVec{x}}\mySet{L}(f_{\myVec{\theta}}, \myVec{x}, \myVec{s})\Big),
\end{equation}
where $\text{sign}(\cdot)$ is the element-wise sign function. Taking the sign of each gradient entry yields uniform perturbation in the direction of the steepest increase in the loss function. 

{\em \Ac{bim}}~\cite{kurakin2018adversarial} is an alternative adversarial attack. Unlike \ac{fgsm}, \ac{bim} does not enforce the perturbation entries to be of magnitude $\epsilon$, allowing  to explore a larger region of the input space. \ac{bim} iteratively refines the perturbation by taking gradient sign steps as in \eqref{eqn:fgsp} followed by their projection onto  $\Delta_{\infty}(\epsilon)$ by clipping, i.e., by iterating for $t=1,2,\ldots$
\begin{equation}
    \myVec{z}_t =  \alpha  \cdot \text{sign}\Big(
  \nabla_{{\myVec{x}}_t}\mySet{L}\Big(f_{\myVec{\theta}}, {\myVec{x}}_t= \big(\myVec{x}+\myVec{\delta}_{t-1}\big), \myVec{s}\Big)\Big),
\end{equation}
and updating $\myVec{\delta}_t = \text{clip}_{\epsilon}( \myVec{z}_t +\myVec{\delta}_{t-1})$, where $\text{clip}_{\epsilon}(\cdot)$ denotes element-wise amplitude clipping  with threshold $\epsilon$, $\myVec{\delta}_0=\myVec{0}$, and $\alpha>0$ is the step-size hyperparameter. 

Additional attacks that are based on \ac{fgsm}, such as,   \ac{nifgsm}~\cite{lin2019nesterov}, as well as alternative methods, e.g., the \ac{cw} attack~\cite{carlini2017towards}, update the perturbation using the model gradients.
\section{Attacking Iterative Optimizers}\label{sec:Attacking} 
While the above attacks were designed for inference rules implemented as parameterized \ac{ml} models, their formulation is invariant of how $f_{\myVec{\theta}}(\cdot)$ is obtained, and one can write the inference rule as $f(\cdot)$. The two properties of \ac{ml} models encapsulated in $f(\cdot)$  that  attacks such as \ac{fgsm} and \ac{bim} exploit are:
\begin{enumerate}[label={P\arabic*}]
    \item \label{itm:differentiable} Differentiability, i.e., one can compute $\nabla_{\myVec{x}}\mySet{L}(f, \myVec{x}, \myVec{s})$.

    \item \label{itm:existence} There exist small perturbations in $\myVec{x}$ that notably affect the output of $f(\cdot)$. 
\end{enumerate}

We next discuss how \ref{itm:differentiable}-\ref{itm:existence} are also satisfied by various  iterative optimizers in Subsection~\ref{ssec:iterative}. Then, we  identify how they are affected by  adversarial examples in Subsection~\ref{ssec:effect}, present how they can be robustified in Subsection~\ref{ssec:unfolding}, and provide a discussion in Subsection~\ref{ssec:discussion}.

%
\subsection{Iterative Optimizer Model}\label{ssec:iterative}  
The term {\em iterative optimizer} refers to iterative methods designed to tackle mathematically formulated  optimization problems. Iterative optimizers can be used to implement inference rules, i.e., mappings of the form $f:\mathcal{X}\mapsto\mathcal{S}$. In such cases, the optimization problem tackled takes the form
\begin{equation}
\label{eqn:OptProb}
    \myVec{s}^{\star} = \mathop{\arg \min}_{\myVec{s}\in \mathcal{S}} \mySet{L}_{\rm op}(\myVec{x},\myVec{s}),
\end{equation}
where $\mySet{L}_{\rm op}(\cdot,\cdot)$ is the {\em objective} function, that often includes some parameters (e.g., regularization coefficients), referred to as the {\em objective hyperparameters}~\cite{shlezinger2022model}. 

\subsubsection{Iterative Methods}
We particularly focus on iterative  methods designed for convex optimization problems\footnote{Convex optimization methods are also widely used  in non-convex setups by, e.g., considering a surrogate convex objective, or by seeking a locally optimal solution~\cite{jain2017non}.}, where the objective function is convex  in $\myVec{s}$ (for all $\myVec{x}$) and $\mathcal{S}$ is convex. These optimizers start from an initial point $\myVec{s}_0$, and on each iteration of index $t$ aim to refine $\myVec{s}_t$ into $\myVec{s}_{t+1}$. We denote the iterative refinement at iterations $t$ as $\myVec{s}_{t+1}=g(\myVec{s}_t;\myVec{x})$, which is typically repeated until convergence. 
These methods are designed to traverse along the loss surface $ \mySet{L}_{\rm op}(\myVec{x}, \cdot)$, starting from $\myVec{s}_0$, until a minima is reached. 
Iterative optimizers often operate as {\em descent methods}, that seek to have $ \mySet{L}_{\rm op}(\myVec{x},\myVec{s}_{t+1}) <  \mySet{L}_{\rm op}(\myVec{x},\myVec{s}_t)$, while searching for $\myVec{s}_{t+1}$ in some bounded environment of $\myVec{s}_t$~\cite{boyd2004convex}.  Their operation introduces additional parameters (e.g., step-sizes), referred to as {\em optimizer hyperparameters}.

\subsubsection{Optimizers as \ac{ml} Models}
Convex iterative optimizers encompass a broad family of different algorithms, which vary based on the specific objective function. A large portion of these algorithms are in fact end-to-end differentiable~\cite{agrawal2021learning}, i.e., hold~\ref{itm:differentiable}. This property is the basis of  {\em deep unfolding}~\cite{monga2021algorithm}, that exploits this differentiability of iterative optimizers to view them as \acp{dnn}. 
In particular, deep unfolding fixes the number of iterations $T$, and then parametrizes the iteration mapping as $g_{\myVec{\theta}_t} (\cdot)$, with the trainable parameters of the $t$th iteration typically being its hyperparameters~\cite{shlezinger2022model}. The resulting iterative optimizer is trained as an \ac{ml} model with parameters $\myVec{\theta}=\{\myVec{\theta}_t\}_{t=1}^T$  and mapping 
\begin{equation}
    f_{\myVec{\theta}}(\myVec{x})=g_{\myVec{\theta}_T} \left(g_{\myVec{\theta}_{T-1}} \left(\cdots g_{\myVec{\theta}_1} \left(\myVec{s}_0 ; \myVec{x}\right) ; \myVec{x}\right) ; \myVec{x}\right).
    \label{eqn:Unfolded}
\end{equation}
The loss function used for training $\myVec{\theta}$ in \eqref{eqn:Unfolded} can potentially differ from the optimization objective  $\mySet{L}_{\rm op}$. A common approach is to train based on using a supervised  loss  $\mySet{L}$ that compares $f_{\myVec{\theta}}(\myVec{x})$ with a desired $\myVec{s}$ based on 
 data $\mySet{D}$, instead of evaluating it via $\mySet{L}_{\rm op}$~\cite{monga2021algorithm}, namely, seek
\begin{equation}
    \myVec{\theta}^{\star} = \mathop{\arg \min}_{\myVec{\theta}} \frac{1}{|\mySet{D}|} \sum_{(\myVec{x},\myVec{s})\in \mySet{D}}\mySet{L}(f_{\myVec{\theta}}, \myVec{x}, \myVec{s}). 
    \label{eqn:SupTrain}
\end{equation}


\subsection{Effects of Attacking Iterative Optimizers}
\label{ssec:effect} 
The fact that many iterative optimizers are end-to-end differentiable, i.e., satisfy \ref{itm:differentiable}, indicates that one can design 
adversarial examples using attacks such as \ac{fgsm} and \ac{bim}.
As mentioned above, iterative optimizers often operate by traversing along a loss surface $ \mySet{L}_{\rm op}(\myVec{x}, \cdot)$, starting from some initial guess $\myVec{s}_0$, and on each iteration take a step based on the loss surface values in some proximity of the current position. The dependence of their operation on the input $\myVec{x}$, which is affected by adversarial examples, is thus encapsulated in the loss surface. Namely, {\em adversarial examples corrupt the loss surface}, having the optimizer choose its steps based on 
\begin{equation}
\label{eqn:AdvSurface}
    \mySet{L}_{\rm adv}(\myVec{x},\myVec{s}) = \mySet{L}_{\rm op}(\myVec{x}+\myVec{\delta}, \myVec{s}).
\end{equation}
As a result, instead of solving \eqref{eqn:OptProb}, the problem which the attacked optimizer ends up tackling becomes
\begin{equation}
    \myVec{s}_{\rm adv}^{\star} = \mathop{\arg \min}_{\myVec{s}\in \mathcal{S}} \mySet{L}_{\rm adv}(\myVec{x},\myVec{s}).
    \label{eqn:AdvProb}
\end{equation}

Comparing~\eqref{eqn:OptProb} with \eqref{eqn:AdvProb} reveals the effect of adversarial attacks on  iterative optimizers: they seek a minor perturbation that notably modifies the minima of the resulting loss surface. For these attacks to be effective with minor  perturbations, \ref{itm:existence} should hold as well. Effective  attacks can be shown to exist in various setups. We next analyze this by focusing separately on optimizers that  run until convergence and ones having a fixed number of iterations. 

\subsubsection{Convergent Optimizers} 
When  tackling a convex problem and allowing to run until convergence, then (assuming  the optimizer hyperparameters allow convergence, e.g., sufficiently small step-sizes~\cite{boyd2004convex}) the non-attacked optimizer outputs $\myVec{s}^{\star}$ in \eqref{eqn:OptProb}, while the attacked one outputs $ \myVec{s}_{\rm adv}^{\star}$ of \eqref{eqn:AdvProb}. 
Therefore,  \ref{itm:existence}, i.e., the existence of an effective adversarial example, depends on the ability of the attacker to make  $ \myVec{s}_{\rm adv}^{\star}$ distinct from $\myVec{s}^{\star}$.  
One such case which can be rigorously analyzed is that of an $\ell_2$ linear data matching objective, stated in the following:
\begin{proposition}
\label{pro:LeastSquares}
Consider an inference rule mapping from $\mySet{X} = \mathbb{R}^n$ into $\mySet{S} = \mathbb{R}^m$, $n > m$, based on the $\ell_2$ objective of the form $\mySet{L}_{\rm op}(\myVec{x},\myVec{s}) = \|\myVec{x}-\myMat{A}\myVec{s}\|^2_2$ for a given $n\times m$ matrix $\myMat{A}$. Then, for any $\epsilon >0$ and for any $p\geq 2$ it holds that  adversarial examples in $\Delta_p(\epsilon)$ can achieve
\begin{equation}
\label{eqn:LeastSquares}
    \|\myVec{s}_{\rm adv}^{\star}  - \myVec{s}^{\star} \|_2^2 \geq \big\| (\myMat{A}^T\myMat{A})^{-1}\myMat{A}^T\big\|_2 \cdot \epsilon,
\end{equation}
    where, when applied to a matrix, $\| \cdot \|_2$ is the $\ell_2$ induced matrix norm, i.e., its maximal singular value. 
\end{proposition}
\begin{IEEEproof}
    The proof is given in Appendix~\ref{app:proof1}. 
\end{IEEEproof}
\smallskip

%

Proposition~\ref{pro:LeastSquares} showcases a simple example where   the optimization objective can be related to adversarial sensitivity. If $\big\| (\myMat{A}^T\myMat{A})^{-1}\myMat{A}^T\big\|_2$ is large, then one can notably shift the optimum point with small yet carefully designed adversarial example.  For the setting in Proposition~\ref{pro:LeastSquares}, the solution to the optimization problem is given in closed-form, not necessitating an iterative procedure to reach the optimum. However, in Section~\ref{sec:sims} we numerically show that sensitivity to adversarial examples is also found in solutions to optimization problems obtained via iterative methods.

\subsubsection{Iteration-Limited Optimizers} 
When an iterative optimizer operates with pre-defined $T$ iterations,  it cannot be guaranteed to recover the optimum, even if the problem is convex. In such cases, performance depends not only on the objective, but also on the solver hyperparameters (e.g., step-sizes), which are often optimized from data in deep unfolding. Accordingly, to characterize adversarial robustness, we recall its relationship with the notion of Lipschitz continuity~\cite{zuhlke2024adversarial}, defined next:
\begin{definition}
\label{def:Lipschitz}
    A mapping $f:\mySet{X}\mapsto\mySet{S}$ is said to be {\em $C$-Lipschitz} in the $\ell_2$ metric if $\forall \myVec{x}_1, \myVec{x}_2 \in \mySet{X}$ it holds that 
    \begin{equation}
        \left\|f(\myVec{x}_1)-f(\myVec{x}_2)\right\|_2 \leq C \cdot \|\myVec{x}_1-\myVec{x}_2\|_2.
    \end{equation}
\end{definition}
Definition~\ref{def:Lipschitz} is clearly tightly coupled with Property~\ref{itm:existence}, as for  $C$-Lipschitz mappings attacked with perturbations in $\Delta_p(\epsilon)$, it holds that the output cannot alter by more than $C\cdot\epsilon$ for every $p \leq 2$. Similarly, for $p > 2$ and $\mySet{X}=\mathbb{R}^n$, the output bound becomes $C \cdot n^{1/2 - 1/p} \epsilon$ by Holder's inequality. In general, the smaller $C$ is, the more  robust the mapping is. 

We next show that for representative families of iteration-limited  optimizers, it holds that $(i)$ the optimizer is $C$-Lipschitz; and $(ii)$ the Lipschitz parameter $C$  depends on the optimizer hyperparameters $\myVec{\theta}$, and thus {\em deep unfolding affects vulnerability to adversarial examples}. Our examples focus on {\em regularized linear least-squares} settings, where $\mySet{X} = \mathbb{R}^n$,  $\mySet{S} = \mathbb{R}^m$, and the optimization objective is 
\begin{equation}
    \label{eqn:Lasso}
    \mySet{L}_{\rm op}(\myVec{x},\myVec{s}) =  \frac{1}{2}\|\myVec{x}-\myMat{A}\myVec{s}\|^2_2 +\rho \cdot \phi(\myVec{s}).
\end{equation}
In \eqref{eqn:Lasso}, $\phi:\mySet{S}\mapsto \mathbb{R}$ is a convex regularizer, while $\rho >0$ and $\myMat{A}\in \mathbb{R}^{n\times m}$ are the objective hyperparameters. This formulation accommodates various problems, including maximum a-posteriori probability estimation in Gaussian noise, and the \ac{lasso} formulation of sparse recovery tasks~\cite{tibshirani1996regression}.

{\bf Example 1: Proximal \ac{gd}} (specialized for \eqref{eqn:Lasso}) refines its estimate of $\myVec{s}$ at each iteration $t$ via
  \begin{equation}
    \label{eqn:ISTA}
         \myVec{s}_{t+1} \!=\! g_{\myVec{\theta}_t}(\myVec{s}_t; \myVec{x})\! =\! {\rm Prox}_{\mu_t\rho_t\cdot \phi}\left( \myMat{M}_t\myVec{s}_t +  \myMat{B}_t \myVec{x}  \right),
    \end{equation}
where $\mu_t$ is the step-size, and   the proximal mapping is defined as ${\rm Prox}_h(\myVec{y})\triangleq \arg\min_{\myVec{z}}\frac{1}{2}\|\myVec{y}-\myVec{z}\|_2^2+h(\myVec{z})$~\cite{parikh2014proximal}. 
The remaining hyperparameters are $\rho_t\equiv \rho$, $\myMat{M}_t \equiv \myMat{I}-\mu_t \myMat{A}^T\myMat{A}$, and $\myMat{B}_t \equiv \mu_t\myMat{A}^T$.
To accommodate unfolded versions of proximal \ac{gd}, we stack the hyperparameters of the $t$th iteration as $\myVec{\theta}_t = \{\mu_t, \rho_t, \myMat{M}_t, \myMat{B}_t\}$. 

\begin{proposition}
\label{thm:PGD}
     Proximal \ac{gd} via \eqref{eqn:ISTA} with $T$ iterations is $C(\myVec{\theta})$-Lipschitz with 
    \begin{equation*}
        C(\myVec{\theta}) = \sum_{t=0}^{T-1} \left(\prod_{j=t+1}^{T-1} \left\|\myMat{M}_j\right\|_2 \right)\mu_t\left\|\myMat{B}_t\right\|_2.
    \end{equation*}
\end{proposition}
\begin{IEEEproof}
    The proof is given in Appendix~\ref{app:proof2}. 
\end{IEEEproof}
\smallskip

{\bf Example 2: \ac{admm}} tackles \eqref{eqn:Lasso} by introducing an  auxiliary variable $\myVec{v}$,  and converting the objective into a constrained  formulation. The latter is then solved via primal-dual alternations between the desired $\myVec{s}$, the auxiliary $\myVec{v}$, and the dual $\myVec{y}$. The resulting iteration $  \myVec{s}_{t+1} = g_{\myVec{\theta}_t}(\myVec{s}_t; \myVec{x})$ is given by
\begin{subequations}
\label{eqn:ADMM}
\begin{align}
      \myVec{s}_{t+1}&\leftarrow \myMat{M}_t(\myVec{v}_t - \myVec{y}_t) +\myMat{B}_t\myVec{x}, \\ 
       \myVec{v}_{t+1}  
	    &\leftarrow   {\rm Prox}_{\frac{1}{2\lambda}\rho_t \cdot \phi}\left(\myVec{s}_{t+1} + \myVec{y}_t\right), \\
       \myVec{y}_{t+1}  
	    &\leftarrow  \myVec{y}_t +\mu_t \left(\myVec{s}_{t+1} -\myVec{v}_{t+1}\right).
\end{align}
\end{subequations}
In \eqref{eqn:ADMM}, $\mu_t$ is the step-size, and the remaining hyperparameters are the scalar $\lambda$ (obtained from the augmented Lagrangian used in \ac{admm} \cite{parikh2014proximal}), the matrices $\myMat{M}_t \equiv \big(\myMat{A}^T\myMat{A}+2\lambda \myMat{I}\big)^{-1}2\lambda$ and $\myMat{B}_t \equiv \big(\myMat{A}^T\myMat{A}+2\lambda \myMat{I}\big)^{-1}\myMat{A}^T$, and the coefficients $\rho_t \equiv \rho$.  As in Example 1, we accommodate unfolded versions of \ac{admm} by stacking the hyperparameters of the $t$th iteration as $\myVec{\theta}_t = \{\mu_t, \frac{1}{2\lambda}\rho_t, \myMat{M}_t, \myMat{B}_t\}$.

\begin{proposition}
\label{thm:ADMM}
Letting $\normtwo{\myMat{B}_t^*}:=\max_{0 \leq i \leq t}{\normtwo{\myMat{B}_i}}$,
    \ac{admm} via \eqref{eqn:ADMM} with $T+1$ iterations is $C(\myVec{\theta})$-Lipschitz with 
    \begin{align*}
        &C(\myVec{\theta}) \!=\! \notag
        \normtwo{\myMat{B}_T^*}\prod_{i=0}^{T}\!\left(\myMat{I}\!+\!2\normtwo{\myMat{M}_{T}}\left(|\mu_{T\!-\!1}|\!+\!1\right)\!\prod_{j=i-1}^{T-2}\!|\mu_j|\right),
    \end{align*}
    using the convention that  product over empty sets equals one.
\end{proposition}
\begin{IEEEproof}
    The proof is given in Appendix~\ref{app:proof3}. 
\end{IEEEproof}
\smallskip

\subsection{Robustness via Unfolding}
\label{ssec:unfolding}
The findings of Examples 1-2 reveal that the hyperparameters of iterative optimizers (of both the objective and the solver) affect their Lipschitz continuity when the number of iterations is fixed. As deep unfolding casts these hyperparameters as trainable weights of an \ac{ml} model, the above indicates that {\em unfolding affects Lipschitz continuity}, which in turn {\em affects adversarial robustness}. 

The effect of training the hyperparameters of an iterative optimizer on its vulnerability to adversarial examples depends on how it is trained. In fact, as we empirically demonstrate in Section~\ref{sec:sims}, when tuning the hyperparameters using conventional training that maximizes performance on a given dataset, i.e., via \eqref{eqn:SupTrain}, the resulting optimizer may actually become more sensitive to adversarial attacks. 

However, the trainability of unfolded optimizers and their effect on Lipschitz continuity can be leveraged to enhance robustness by solely altering its hyperparameters via {\em adversarial training}. For instance, this can be achieved by training $\myVec{\theta}$ to approach~\cite{madry2017towards}
\begin{equation}
    \myVec{\theta}^{\star}_{\rm adv} = \mathop{\arg \min}_{\myVec{\theta}} \mathop{\max}_{\myVec{\delta} \in \Delta_p(\epsilon)} \frac{1}{|\mySet{D}|} \sum_{(\myVec{x},\myVec{s})\in \mySet{D}}\mySet{L}(f_{\myVec{\theta}}, \myVec{x} +\myVec{\delta} , \myVec{s}). 
    \label{eqn:advTrain}
\end{equation}
Such adversarial training of unfolded optimizers is systematically shown to enhance adversarial robustness in Section~\ref{sec:sims}. It is emphasized that leveraging unfolding for adversarial robustness does not modify the overall iterative algorithm, only its hyperparameters. This is as opposed to seeking an alternative formulation for the objective, e.g., minimax~\cite{thekumparampil2019efficient}, that modifies the overall iterative algorithm and not just its parameters. Instead, \eqref{eqn:advTrain} merely alters the training procedure via \eqref{eqn:advTrain}.




 
\subsection{Discussion}
\label{ssec:discussion}

The above analysis, along with the numerical study reported in Section~\ref{sec:sims}, reveal that iterative optimizers are often sensitive to adversarial examples, a property typically considered to be a peculiarity of \ac{ml} models, and particularly \acp{dnn}~\cite{szegedy2013intriguing}. In fact, \ac{ml} training mechanisms were in some cases  evaluated based on their ability to achieve similar adversarial robustness as iterative optimizers~\cite{genzel2022solving}.
Our analysis reveals a new form of a threat to iterative optimizers, which are widely used in signal processing and communications~\cite{luo2006introduction}. For instance, end-to-end differentiable iterative optimizers are common in control~\cite{agrawal2020learning}, beamforming~\cite{lavi2023learn}, and wireless receiver processing~\cite{yang2015fifty}. For the latter, adversarial vulnerability indicates the potential of new risks in the form of low-power hard-to-detect jamming mechanisms. 

While iterative optimizers can be sensitive to adversarial examples, their interpretable nature allows to directly trace the effect of adversarial examples to the loss surface considered. This indicates that iterative optimizers can possibly cope with adversarial attacks by proper selection of the objective used for the considered task. A natural approach to boost adversarial robustness is thus to adopt a robust optimization framework, e.g., utilizing a minimax objective~\cite{thekumparampil2019efficient}. An alternative approach is based on identifying that while adversarial attacks affect the objective, they are obtained from the inference rule, i.e., the iterative solver. For instance, in Section~\ref{sec:sims} we demonstrate that two different optimizers tackling the same optimization problem demonstrate different levels of adversarial sensitivity. Consequently, one can  affect  sensitivity to some extent by opting a specific solver and via its hyperparameters, e.g., the maximal number of iterations, step sizes, and regularization coefficients. This capability is leveraged by adversarial-aware unfolding approach, detailed in Subsection~\ref{ssec:unfolding}. 

\section{Experimental Study}
\label{sec:sims}
In this section we numerically investigate the vulnerability  to adversarial examples of   iterative optimizers\footnote{The source code used in this section is available online at \url{https://github.com/eladgsofer/adversarial_updated}.}.
Our experimental study is divided into two parts: The first part, detailed in Subsection~\ref{subsec:simsOpt}, numerically demonstrates the vulnerability to adversarial examples of several representative iterative optimizers for three case studies in sparse recovery, \ac{rpca}, and hybrid beamforming. The second part, presented in Subsection~\ref{subsec:simsUnfold}, investigates how  unfolding affects this sensitivity. 

\subsection{Adversarial Sensitivity of Iterative Optimizers}
\label{subsec:simsOpt}

\subsubsection{Case Study I: Sparse Recovery}
We first simulate a \ac{cs} setup. Here, $\myVec{s}$ is a $5$-sparse  vector in $\mathbb{R}^{1500}$, whose non-zero entries are generated from a zero-mean Gaussian distribution with standard deviation of $0.5$. The input $\myVec{x}$ is a $256\times  1$ vector obtained as $\myVec{x}=\myMat{H}\myVec{s}+\myVec{w}$, where $\myVec{w}$ and $\myMat{H}$ are comprised of i.i.d. Gaussian entries with standard deviations of $0.01$ and $1$, respectively. 
The recovery of $\myVec{s}$ from $\myVec{x}$ is formulated via the \ac{lasso}~\cite{tibshirani1996regression}, resulting in the convex objective in \eqref{eqn:Lasso}, with $\phi(\cdot)$ being the $\ell_1$ norm, and with $\rho = 0.01$.

\begin{figure}
    \centering
    \includegraphics[width=\linewidth]{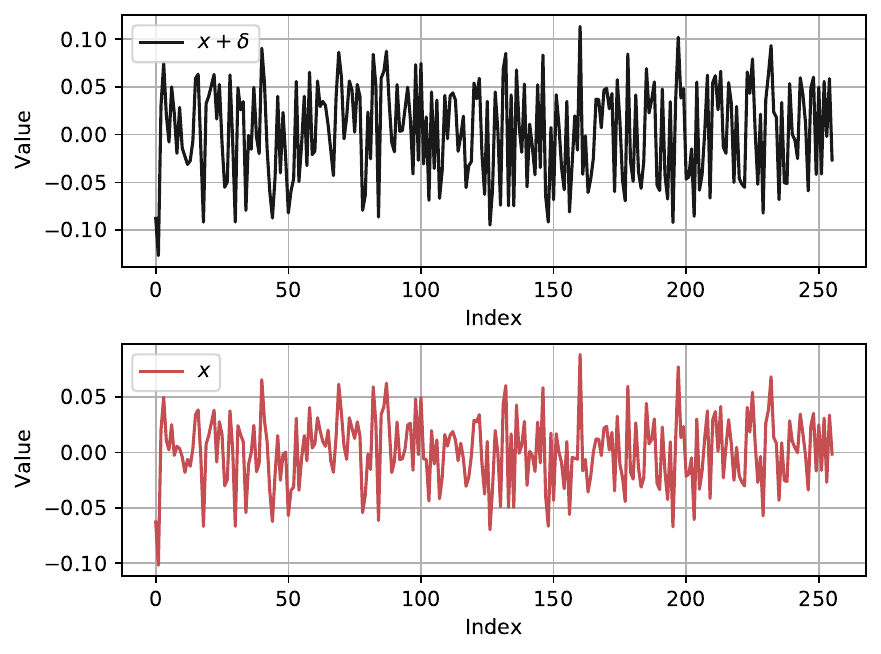}
    \vspace{-0.3cm}
    \caption{Illustration of $\myVec{x}$ and its associated  $\myVec{x}+\myVec{\delta}$ for $\epsilon=0.025$.}
    \label{fig:ObsCombined}
\end{figure}

\begin{figure}
    \centering
    \includegraphics[width=\linewidth]{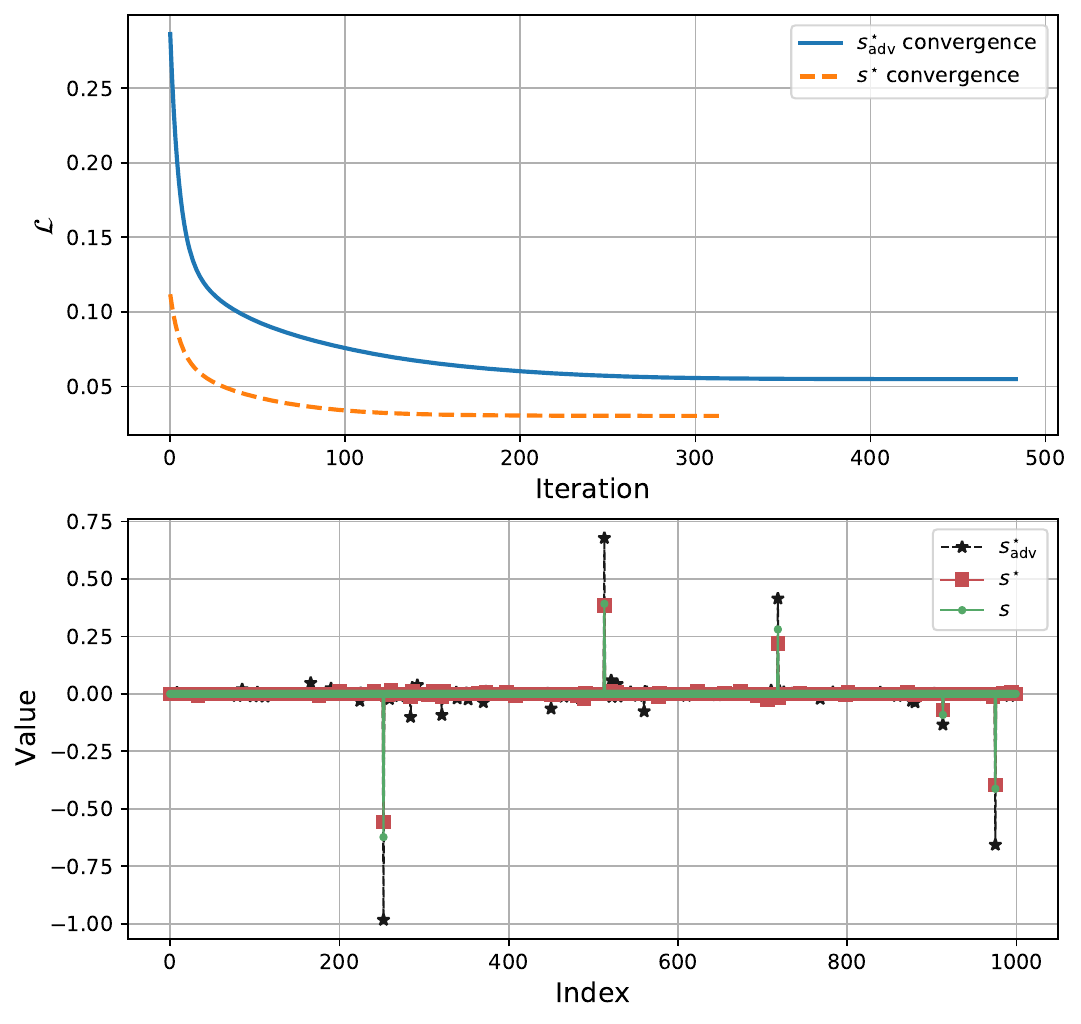}
    \vspace{-0.3cm}
    \caption{ISTA $\myVec{x}+\myVec{\delta}$ and $\myVec{x}$ convergence for $\epsilon=0.025$.}
    \label{fig:ObsConv}
\end{figure}

To visualize the effect of adversarial examples, we adopt the \ac{ista}~\cite{daubechies2004iterative}, which specializes proximal \ac{gd} for the \ac{lasso} objective. We consider a single realization of $(\myVec{x},\myVec{s})$, and perturb the input with \ac{bim} attack (applied to  \ac{ista}) of radius $\epsilon = 0.025$. The resulting input and its perturbed version are depicted in Fig.~\ref{fig:ObsCombined}, showing the stark similarity and near-indistinguishability between $\myVec{x}$ and $\myVec{x}+\myVec{\delta}$. Nonetheless, the adversarial example dramatically affects the operation of the iterative optimizer, resulting in slower convergence and the estimate of a vector $\myVec{s}_{\rm adv}^\star$ that is substantially different from $\myVec{s}$, as illustrated in Fig.~\ref{fig:ObsConv}.

\begin{figure} 
    \centering
    \includegraphics[width=0.9\linewidth]{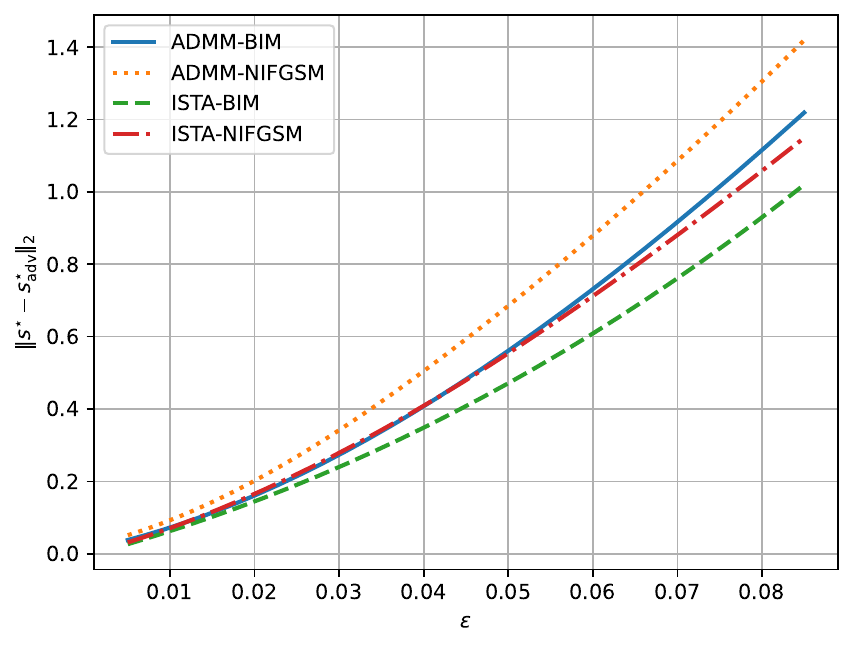}
    \caption{Distortion ${\| \myVec{s}^{\star} - \myVec{s}_{\rm adv}^{\star} \|}_2$ versus attack radius $\epsilon$.}
    \label{fig:CSCombined}
\end{figure}

To illustrate that the observed sensitivity is not unique to a specific optimizer and attack, we employ both \ac{bim} and NIFGSM to generate attacks in $\Delta_{\infty}(\epsilon)$ with $\epsilon \in [0.005,0.085]$, targeting both \ac{ista} and \ac{admm} applied to the \ac{lasso} objective. In Fig.~\ref{fig:CSCombined} we report the distortion induced by this attack, comparing the gaps $\|\myVec{s}^\star-\myVec{s}_{\rm adv}^\star\|_2$ averaged over ${1200}$ Monte Carlo trials. We observe in Fig.~\ref{fig:CSCombined} that both algorithms are affected by both adversarial attacks. Yet, we also observe some minor differences in the induced distortion. This  indicates that while the effect of adversarial examples on iterative optimizers is reflected in the objective, the fact that the attack  (e.g., \ac{bim}) is based on the specific iterative optimizer results in some variations in the overall distortion. 

\subsubsection{Case Study II: \ac{rpca}}
While the first study considered a convex optimization setting, with synthetic data, our next case study considers a non-convex optimizer applied with real world data. We focus on the \ac{rpca} task, which deals with the separation of an image $\myMat{M}$ into a low rank matrix $\myMat{L}$ and a sparse matrix $\myMat{S}$~\cite{candes2011robust}.
While \ac{rpca} is often relaxed into a convex formulation, we focus on its direct formulation, given by 
\begin{align}    
\label{eqn:rpca}
    &\mySet{L}_{\rm op}\left(\myVec{x}=\myMat{M},\myVec{s}=\big[\myMat{L}, \myMat{S}\big]\right) = \left\|\myMat{M}-\myMat{L} - \myMat{S} \right\|_F  \\
    &\text{subject to } {\rm rank}(\myMat{L})\le r, ~\text{and}~~\|\myMat{S}\|_0\le k,\notag
\end{align}
where $k$ and $r$ are hyperparameters. 

\begin{figure*}
    \begin{subfigure}[b]{\columnwidth}
    \centering
    \includegraphics[width=\linewidth]{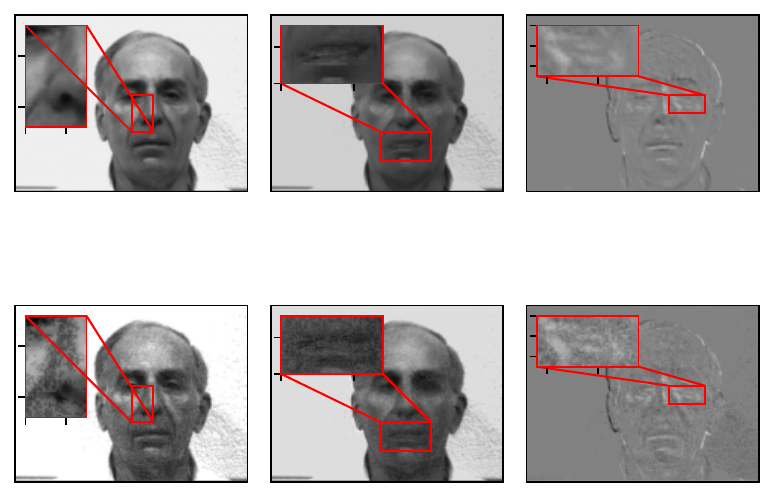}   
    \caption{Subject 1 (using \ac{bim} attack).} 
    \end{subfigure}
        \begin{subfigure}[b]{\columnwidth}
    \centering
    \includegraphics[width=\linewidth]{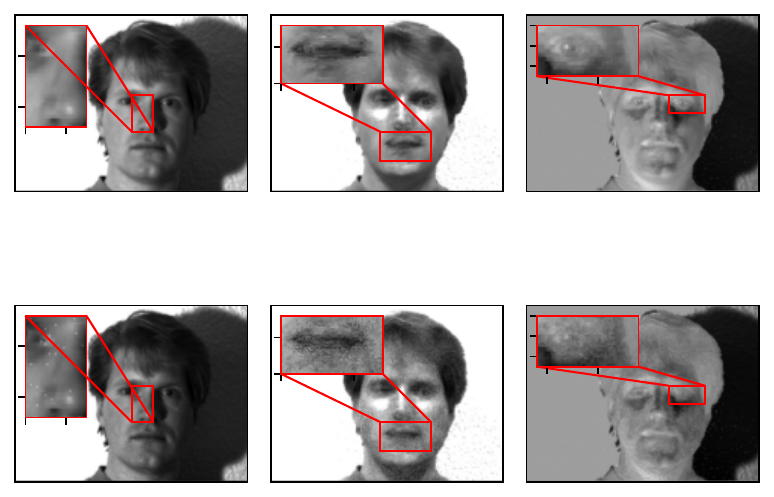}      
    \caption{Subject 2 (using \ac{cw} attack).} 
    \end{subfigure}
    \vspace{0.4cm}
    \caption{Two example images (left column) decomposed into an expressionless low-rank component (middle column) and an expression sparse component (right) column using AccAltProj with clean input (upper row) and perturbed input (lower row).}
    \label{fig:attack_RPCA_Yale}
\end{figure*}

A candidate optimizer for tackling the non-convex objective is the accelerated version of the alternating projections method of \cite{netrapalli2014non}, termed AccAltProj.  We next apply AccAltProj to the Yale Face Database~\cite{belhumeur1997eigenfaces}, where it is known to divide a set of 11 image of the same person into an expressionless image and a set of expression images~\cite{tan2023deep}. We attack AccAltProj  using \ac{bim} with $\epsilon=0.07$ and \ac{cw} with trade-off parameter of $8.8*10^5$, 
visualizing representative subjects in Fig.~\ref{fig:attack_RPCA_Yale}. The results showcase the effect of the adversarial attack on the output of the iterative optimizer. While the differences in the input are hardly visible, the attack results in AccAltProj yielding images in which  key features, such as mouth and eyes, are blurred, and the effects of the attack are clearly visible.

\begin{figure}
    \centering
    \includegraphics[width=0.9\linewidth]{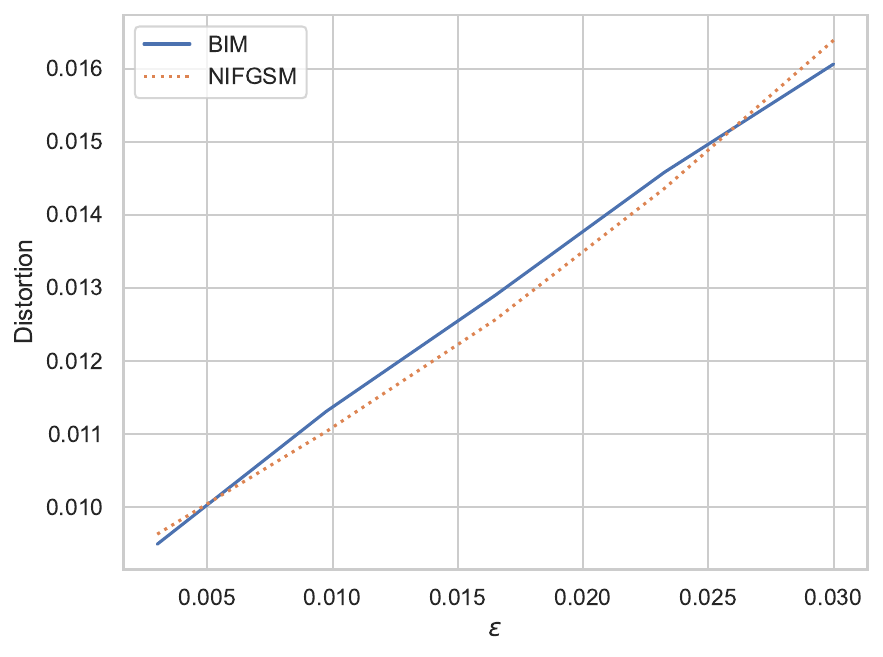}
    \caption{AccAltProj - Distortion versus $\epsilon$}
    \label{fig:rpca_simulated_data}
\end{figure}

 So far we demonstrated the ability to attack AccAltProj which performs RPCA while optimizing Eq.~\eqref{eqn:rpca}. We used the Yale dataset, providing visual examples of degrading AccAltProj performance while injecting adversarial examples as input. To generalize our findings, we examine AccAltProj sensitivity upon simulated data, averaging the distortion upon $N=100$ Monte Carlo trials for each $\epsilon \in [0.003, 0.03]$.
We implement two representative attacks--\ac{bim} and \ac{nifgsm}--and report the results in Fig.~\ref{fig:rpca_simulated_data}. The findings  in Fig.~\ref{fig:rpca_simulated_data} are fully in line with those reported  for the convex optimizers of Case Study I. It is clearly evident that despite the non convexity arising in Case Study II, the sensitivity and vulnerability of the iterative optimizer to the different attacks still holds.

\subsubsection{Case Study III: Hybrid Beamforming}

Our third case study considers a hybrid beamforming setup, where iterative algorithms are commonly employed~\cite{shlezinger2023ai}. Such settings involve the beamforming of an outgoing communication signal using digital and analog precoders, aiming to maximize the communication rate. Specifically, we consider the transmission of a wideband signal with $B=16$ frequency bins intended to $N=4$ receivers  by a transmitter with $12$ antennas and $10$ RF chains. 
In this case, the input $\myVec{x}$ is the set of $4\times 12$ complex matrices $\{\myMat{H}_b\}_{b=1}^{B}$, and the optimization variables $\myVec{s}$ are the  $12\times 10$ analog precoder matrix $\myMat{W}_a$ and the set of $10 \times 4$ digital precoder matrices $\{\myMat{W}_{d,b}\}_{b=1}^{B}$. The achievable rate objective is
\begin{align*}    
    &\mySet{L}_{\rm op}\left(\myVec{x}=\{\myMat{H}_b\},\myVec{s}=\big[\myMat{W}_a,  \{\myMat{W}_{d,b}\}\big]\right) \notag \\
    &= \frac{-1}{B} \sum_{b=1}^{B} \log \left|\myMat{I}_N \!+\! \frac{1}{N \sigma^2} \myMat{H}_b\myMat{W}_a\myMat{W}_{d,b} \myMat{W}_{d,b}^H\myMat{W}_a^H\myMat{H}_b^H\right|,
\end{align*}
where $\sigma^2$ denotes the noise variance, set here to unity. 

\begin{figure} 
    \centering
    \includegraphics[width=0.9\linewidth]{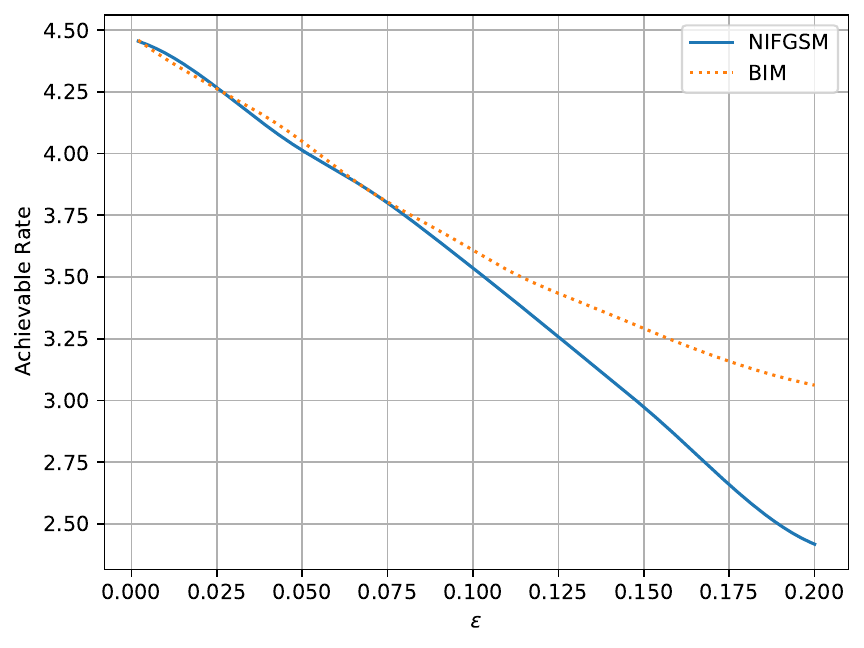}
    \caption{Achievable rate versus attack radius $\epsilon$.}
    \label{fig:HBF}
\end{figure}

We apply the \ac{pgd} iterative algorithm (not to be confused with an adversarial attack  of a similar name) described in \cite{lavi2023learn} to optimize $\myVec{s}$ based on the achievable rate objective. The channel matrices $\{\myMat{W}_{d,b}\}$ are obtained from the QuaDRiGa   channel model~\cite{jaeckel2014quadriga}, using $100$ realizations. We evaluate the sensitivity of \ac{pgd} by employing \ac{bim} and \ac{nifgsm} with attack radius of $\epsilon\in\{0.002,0.2\}$, corresponding to maximal change in the channel norm of $0.5\%$ (for $\epsilon=0.002$) to $32.9\%$ (for $\epsilon=0.2$). Since the gradients of the rate objective were shown to be relatively large, we replaced the sign gradient of \ac{bim}  with a standard gradient. 

The effect of adversarial examples on \ac{pgd} are reported in Fig.~\ref{fig:HBF}. There, we observe that the carefully designed perturbations can notably degrade the rate. For comparison, in order to achieve the degradation induced by a perturbation with $\epsilon=0.15$ using conventional jamming, one would have to increase the noise power by over $50\%$. These results indicate a possible threat of adversarial examples on iterative optimizers, enabling new forms of sophisticated jamming.

\subsection{Robustness of Unfolded Optimizers}
\label{subsec:simsUnfold}
We proceed to evaluate the ability to alleviate sensitivity to adversarial attacks of iterative optimizers by casting them as \ac{ml} models via deep unfolding. To that aim, we focus on Case Study I, for which we employ \ac{ista} as a suitable iterative optimizer. Here, we consider its unfolding into \ac{lista}~\cite{gregor2010learning}, which follows the parameterization introduced in \eqref{eqn:ISTA}, while fixing $T={5}$ iterations. For training the resulting \ac{ml} model, we employed $(i)$ standard empirical risk minimization via \eqref{eqn:SupTrain}, which we term {\em LISTA}; and $(ii)$ adversarial training using \eqref{eqn:advTrain}, coined {\em Robust LISTA}. We analyze the effect of adversarial training compared to standard learning of the iterative optimizer in three aspects: robustness to adversarial examples, performance on non-attacked examples, and Lipschitz continuity bound.

\begin{figure}
    \centering
    \includegraphics[width=0.9\linewidth]{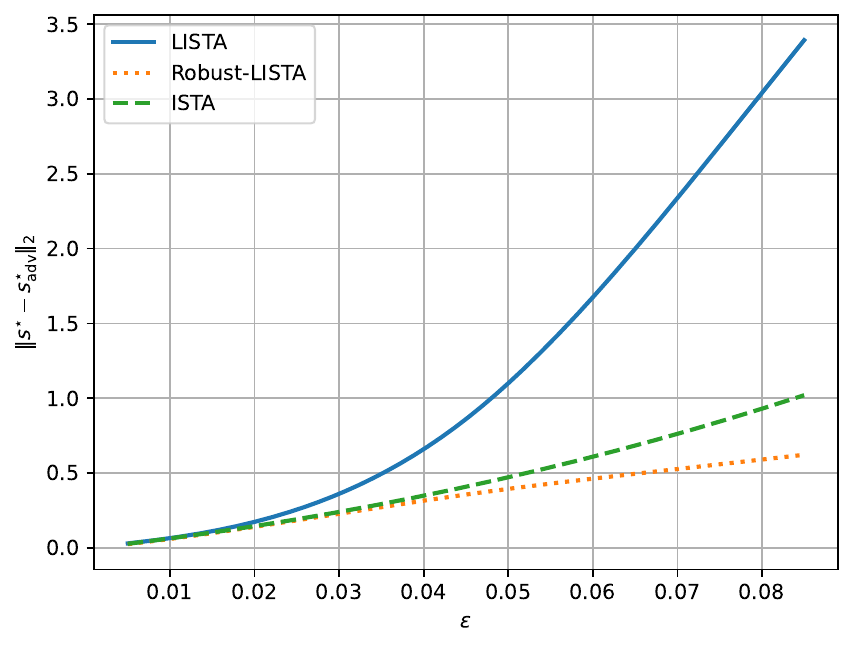}
    \caption{Distortion versus $\epsilon$, perturbed input.}
    \label{fig:defense_adv_data_LISTA}
\end{figure}

In Fig.~\ref{fig:defense_adv_data_LISTA} we report the distortions obtained by the considered optimizers compared to conventional \ac{ista} (which runs until convergence) when the input is perturbed using \ac{bim}. We observe that standard unfolding can in fact increase vulnerability to adversarial attacks, as LISTA is shown to be most sensitive. However, when trained in an adversarial manner, unfolding is in fact shown to alleviate adversarial sensitivity. 

\begin{figure}
    \centering
    \includegraphics[width=0.9\linewidth]{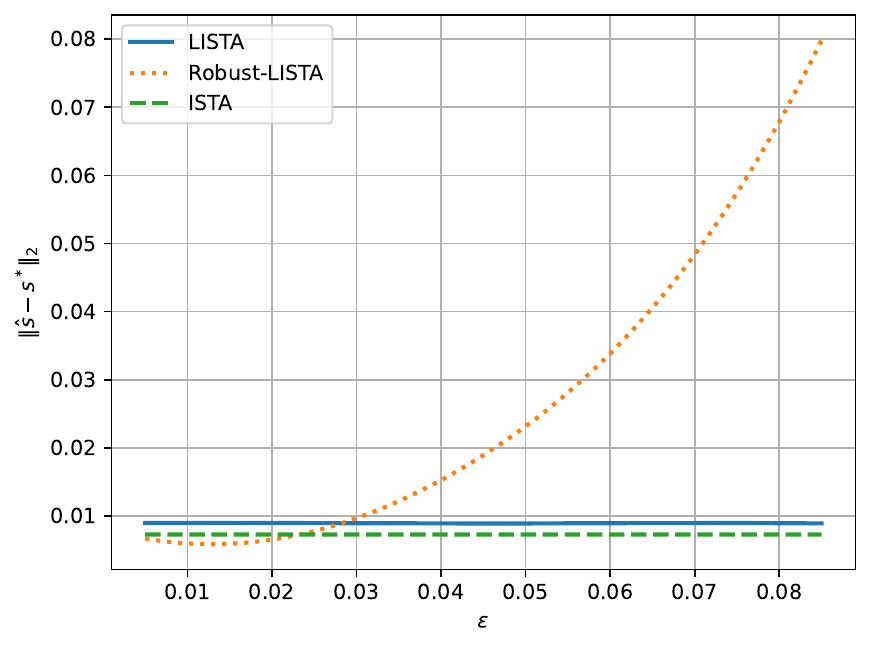}
    \caption{Distortion versus $\epsilon$, clean input.}
    \label{fig:defense_clean_data_LISTA}
\end{figure}

\begin{figure}
    \centering
    \includegraphics[width=0.9\linewidth]{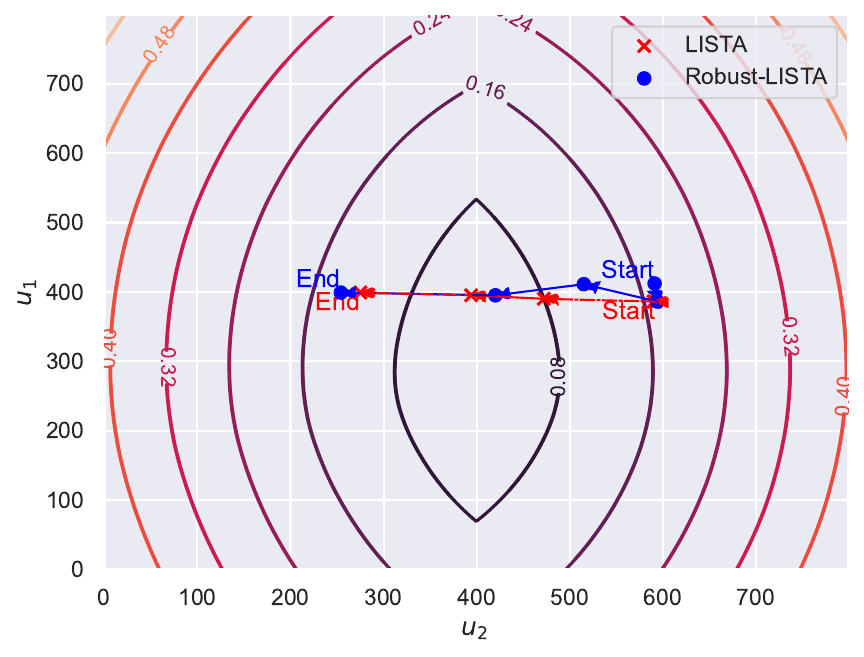}
    \caption{Optimization trajectory on 2D projected objective.}
    \label{fig:LISTA_trajectories}
\end{figure}

The robustness induced into iterative optimizers via adversarial training comes at the cost of a minor degradation in performance when applied to clean (non-attacked) inputs. This is observed in Fig.~\ref{fig:defense_clean_data_LISTA}, where we compare robust LISTA (trained for different levels of attack $\epsilon$) to LISTA and ISTA. There, we note that for small $\epsilon$, adversarial training achieves similar performance on clean data as the convergent ISTA does (and can even  yield better performance). When learning to cope with more pronounced attacks (larger $\epsilon$), adversarial training leads to some performance degradation, though, comparing the y-axis of Figs.~\ref{fig:defense_adv_data_LISTA} and \ref{fig:defense_clean_data_LISTA}, we observe that the degradation in performance is smaller by an order of magnitude compared to that induced by adversarial attacks. The fact that these variations due to adversarial training are relatively minor is also observed in Fig.~\ref{fig:LISTA_trajectories}, where we visualize the overall trajectory of LISTA and robust LISTA (trained with $\epsilon=0.025$) over a 2D projection of the LASSO objective (obtained using the projection method of \cite{li2018visualizing}). There, we see that the optimization trajectories are almost identical, thus adversarial training only mildly affects the optimization procedure when applied to clean inputs. 

\begin{figure*}
	\centering
	\begin{subfigure}[b]{0.45\linewidth}
		\centering
		\includegraphics[width=\linewidth]{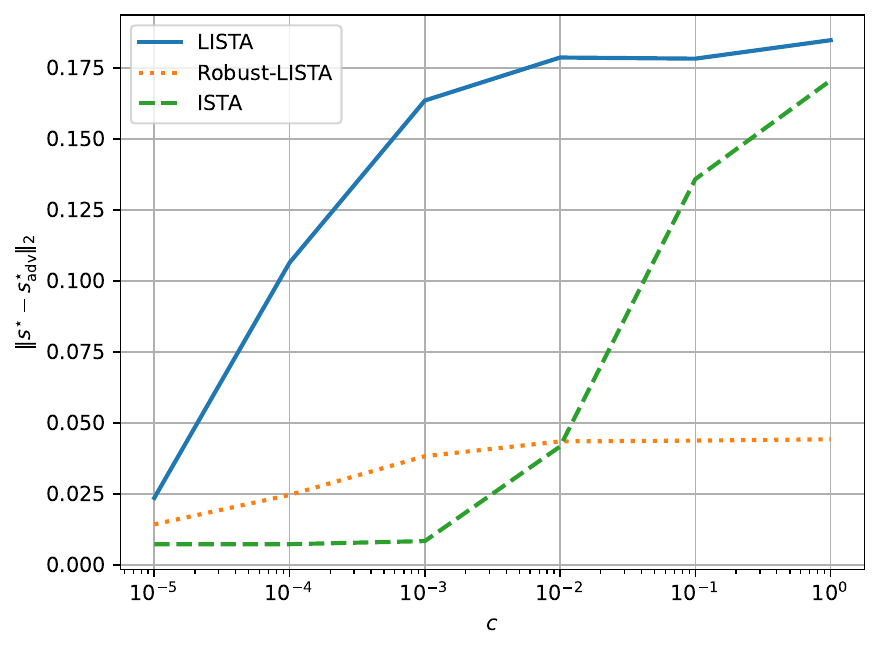}
		\caption{Distortion vs $c$, adversarial input}
		\label{fig:defense_adv_data_LISTA_CW}
	\end{subfigure}
	\hspace{0.05\linewidth} 
	\begin{subfigure}[b]{0.45\linewidth}
		\centering
		\includegraphics[width=\linewidth]{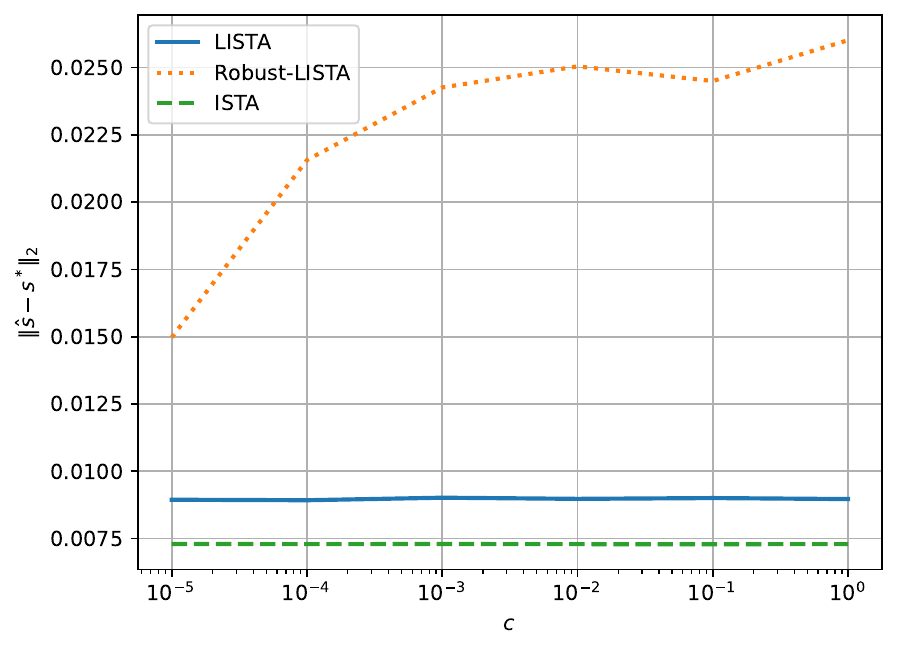}
		\caption{Distortion vs c, clean input}
		\label{fig:defense_clean_data_LISTA_CW}
	\end{subfigure}
	\vspace{0.4cm}
	\caption{Comparison of results using \ac{cw} attacks.}
	\label{fig:LISTA_CW}
\end{figure*}

In  Fig.~\ref{fig:LISTA_CW} we report the sensitivity to \ac{cw} attack. We examine and compare the adversarial sensitivity of ISTA, LISTA, and Robust-LISTA, which was trained with CW perturbations during its adversarial training procedure.
For \ac{cw}, the main hyperparameter controlling the magnitude of the perturbation is the tradeoff parameter $c$. 
We test a range of \( c \) values \(\{0.00001, 0.0001, 0.001, 0.01, 0.1, 1\}\) and average distortion across 1200 Monte Carlo trials for each \( c \).

The findings reported in Fig.~\ref{fig:defense_adv_data_LISTA_CW} preserve the same trend observed in  Fig.~\ref{fig:defense_adv_data_LISTA}, namely, that Robust-LISTA systematically achieves a notably improved robustness to adversarial attack. We also see in Fig.~\ref{fig:defense_clean_data_LISTA_CW} that the gained robustness comes at a relative marginal distortion when the data is not perturbed. 

\begin{figure*}
	\centering
	\begin{subfigure}[b]{0.45\linewidth}
		\centering
		\includegraphics[width=\linewidth]{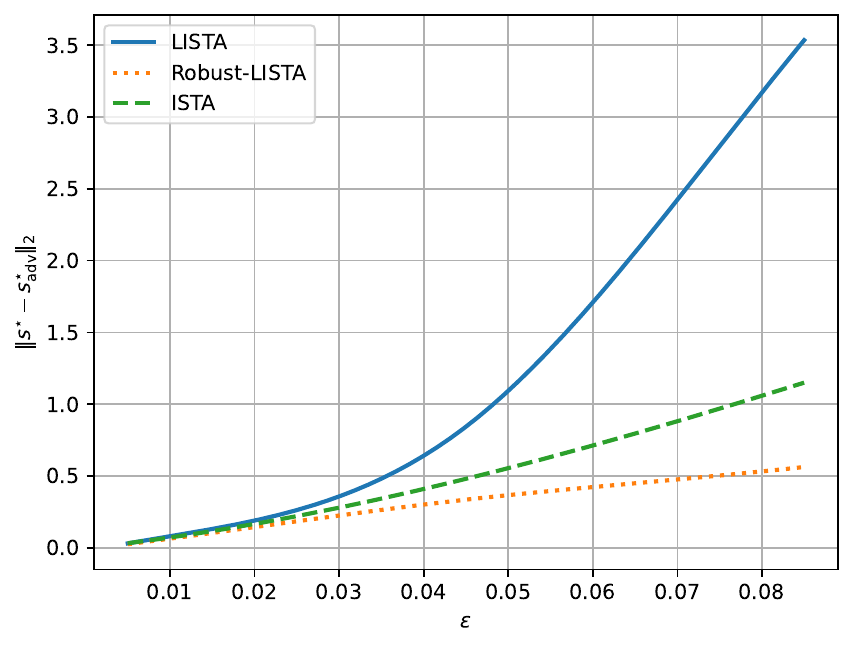}
		\caption{Distortion vs $\epsilon$, adversarial input}
		\label{fig:defense_adv_data_LISTA_NIFGSM}
	\end{subfigure}
	\hspace{0.05\linewidth} 
	\begin{subfigure}[b]{0.45\linewidth}
		\centering
		\includegraphics[width=\linewidth]{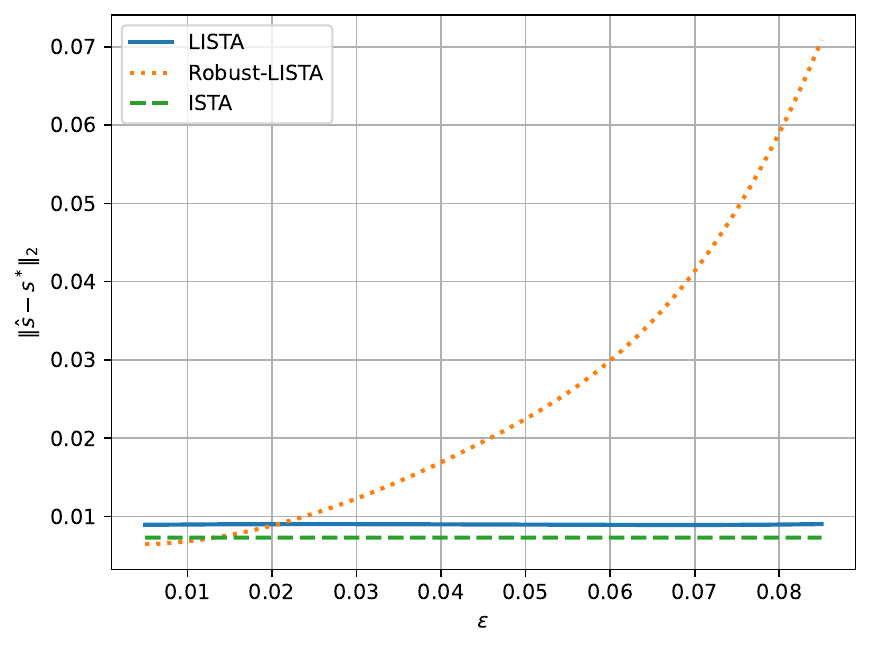}
		\caption{Distortion vs $\epsilon$, clean input}
		\label{fig:defense_clean_data_LISTA_NIFGSM}
	\end{subfigure}
	\vspace{0.4cm}
	\caption{Comparison of clean and adversarial data for LISTA using NIFGSM.}
	\label{fig:LISTA_NIFGSM}
\end{figure*}


Next, we strengthen our findings by repeating the same experiment with a different type of attack, \ac{nifgsm} selecting an \(\epsilon\) range of $[0.005, 0.085]$. We use NIFGSM both for attacking LISTA, ISTA, Robust-LISTA  and for training Robust-LISTA as shown in Fig.~\ref{fig:LISTA_NIFGSM}. Specifically,  in Fig.~\ref{fig:defense_adv_data_LISTA_NIFGSM} we report the distortion compared to the  attack magnitude ($\epsilon$), while in Fig.~\ref{fig:defense_clean_data_LISTA_NIFGSM} we show the performance achieved with clean data compared to a trained Robust-LISTA. Again, we observe that adversarial-aware unfolding yields improved robustness at the cost of relatively minor degradation. The results are fully aligned with those observed for the BIM attack, emphasizing the replicability and the generality of our findings.

\begin{figure*}
	\centering
	\begin{subfigure}[b]{0.45\linewidth}
		\centering
		\includegraphics[width=\linewidth]{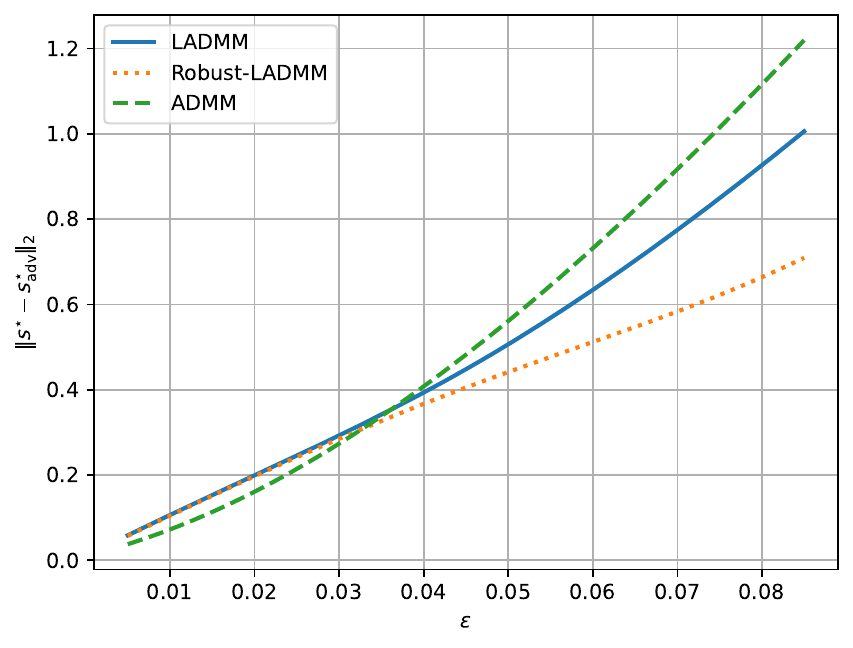}
		\caption{Distortion vs $\epsilon$, clean input}
		\label{fig:defense_adv_data_LADMM_BIM}
	\end{subfigure}
	\hspace{0.05\linewidth} 
	\begin{subfigure}[b]{0.45\linewidth}
		\centering
		\includegraphics[width=\linewidth]{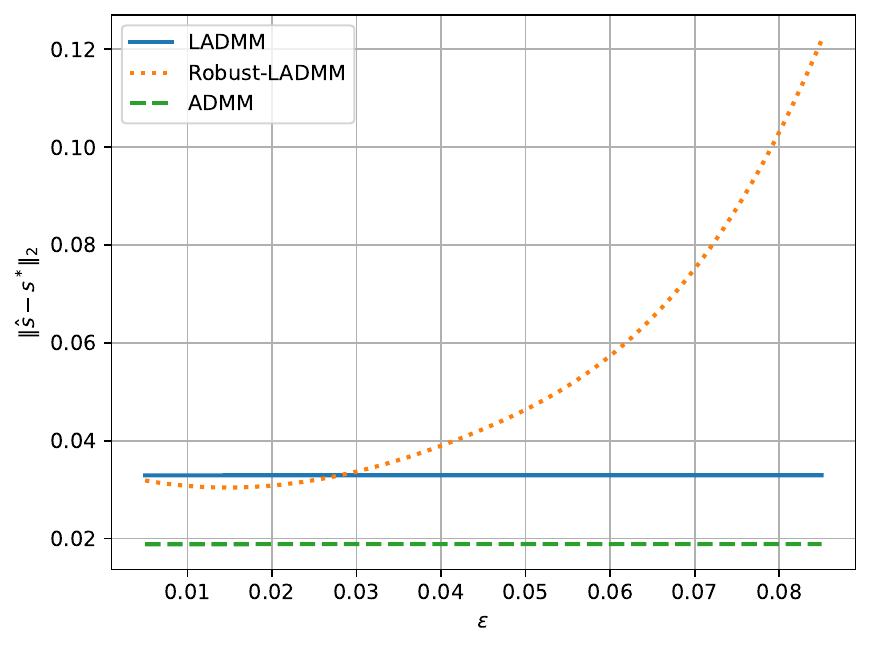}
		\caption{Distortion vs $\epsilon$, clean input}
		\label{fig:defense_clean_data_LADMM_BIM}
	\end{subfigure}
	\vspace{0.4cm}
	\caption{Comparison of clean and adversarial data for LADMM using BIM.}
	\label{fig:LADMM_BIM}
\end{figure*}

\begin{figure*}
	\centering
	\begin{subfigure}[b]{0.45\linewidth}
		\centering
		\includegraphics[width=\linewidth]{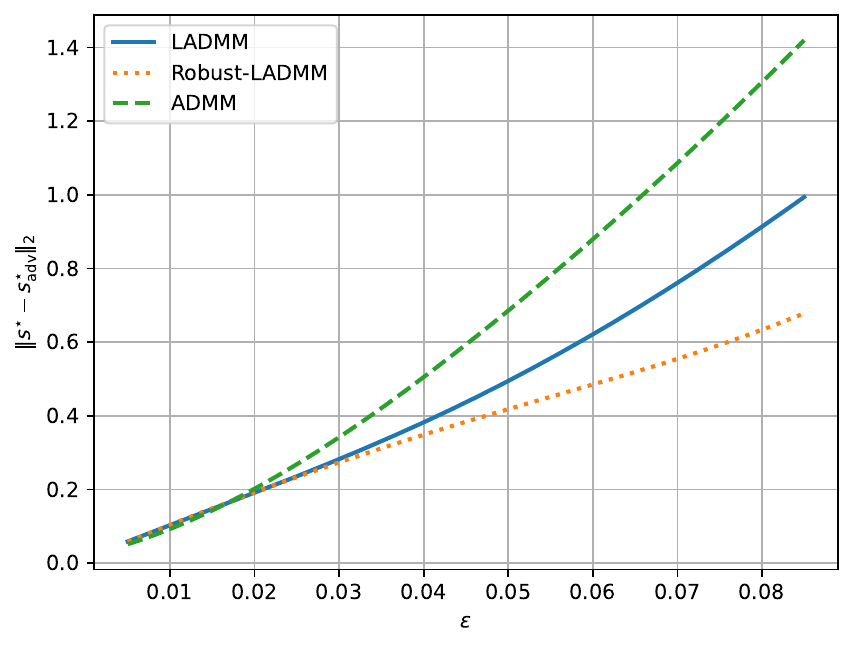}
		\caption{Distortion vs $\epsilon$, clean input}
		\label{fig:defense_adv_data_LADMM_NIFGSM}
	\end{subfigure}
	\hspace{0.05\linewidth} 
	\begin{subfigure}[b]{0.45\linewidth}
		\centering
		\includegraphics[width=\linewidth]{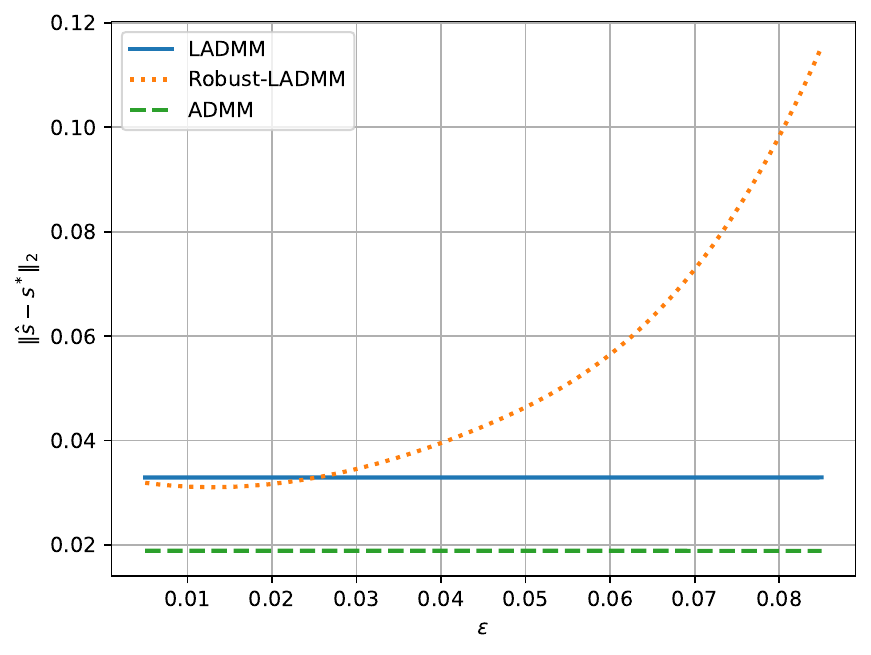}
		\caption{Distortion vs $\epsilon$, clean input}
		\label{fig:defense_clean_data_LADMM_NIFGSM}
	\end{subfigure}
	\vspace{0.4cm}
	\caption{Comparison of clean and adversarial data for LADMM using NIFGSM.}
	\label{fig:LADMM_NIFGSM}
\end{figure*}

To demonstrate that the gains of adversarial-aware unfolding are not unique to \ac{lista}, we next explore the sensitivity and robustness of a different solver algorithm, ADMM. We also consider an unfolding of \ac{admm} in which the hyperparameters of the solver are cast as trainable parameters with $T=6$ iteration. The unfolded versions is termed {\em LADMM}.
As we clearly see in  Fig.~\ref{fig:LADMM_BIM}  the recurring patterns previously identified: the robust model, Robust-LADMM, demonstrates the highest adversarial robustness but exhibits the lowest performance on clean data. 
Additionally, in Fig.~\ref{fig:defense_adv_data_LADMM_BIM}, LADMM proves to be more resilient than vanilla ADMM. in contrast to our findings in  Fig.~\ref{fig:defense_adv_data_LISTA}, where LISTA was more sensitive to adversarial input compared to vanilla ISTA.
To further validate our findings, we repeat this exact same experiment with NIFSGM attack, witnessing at Fig.~\ref{fig:LADMM_NIFGSM} the same patterns with minor differences in the performance comparing to our findings at Fig.~\ref{fig:LADMM_BIM}.

\begin{figure}
    \centering
    \includegraphics[width=0.9\linewidth]{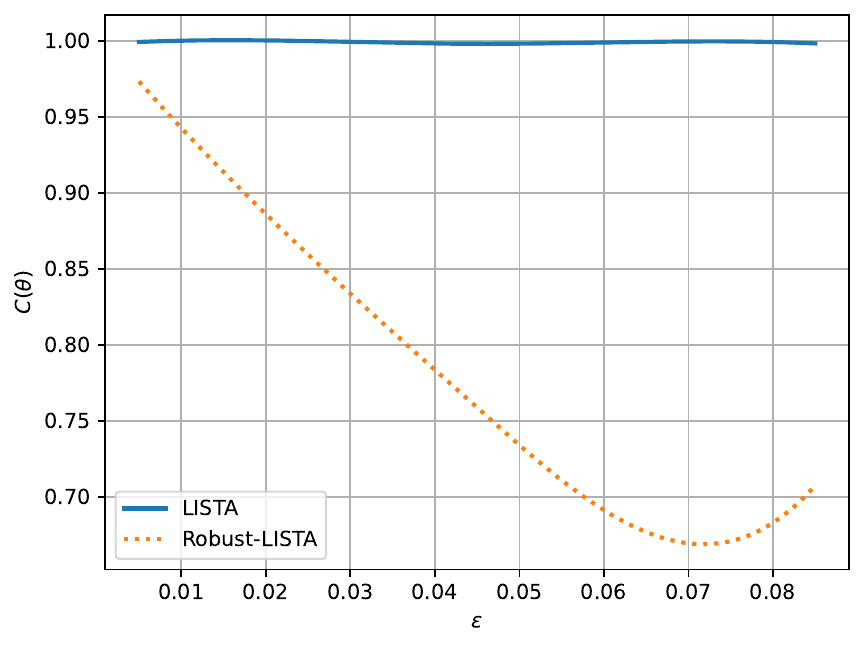}
    \caption{Normalized Lipschitz continuity bound.}
    \label{fig:bound_graph}
\end{figure}

We conclude by empirically demonstrating the effect of adversarial training on the Lipschitz continuity constant of the iterative optimizer. Specifically, in Proposition~\ref{thm:PGD} we showed that the Lipschitz constant of proximal \ac{gd} (which accommodates \ac{ista}) depends on the parameters learned in unfolding. In Fig.~\ref{fig:bound_graph} we report the Lipschitz constant obtained with adversarial training (normalized by that of standard LISTA). It is shown that adversarial training of iterative optimizers, although not explicitly encouraged to enhance continuity, indeed results in a smaller Lipschitz constant. Thus, the ability to affect continuity via unfolding is shown to translate into improved (smaller) Lipschitz constant via adversarial training, which in turn directly reflects on adversarial sensitivity, as observed above.

\section{Conclusions}
\label{sec:conclusions}
In this work we studied the sensitivity to adversarial examples of iterative optimizers, traditionally considered to be a property of \ac{ml} models. We identified that convex optimization methods are often amenable to adversarial attacks being end-to-end differentiable and in some cases sensitive to small variations in their input. Specifically, these variations are reflected in a modified objective,  leading to a deviation in the minima. We revealed that the ability to cast iteration-limited optimizers as \ac{ml} models via deep unfolding can be leveraged to affect adversarial sensitivity. This was shown both theoretically, by characterizing the relationship between the learned optimizer hyperaparameters and Lipschitz constant for representative optimizers, as well as via empirical observations.


 \appendix
\section{Appendix}
\numberwithin{lemma}{subsection} 
\numberwithin{corollary}{subsection} 
\numberwithin{remark}{subsection} 
\numberwithin{equation}{subsection}	

\subsection{Proof of Proposition~\ref{pro:LeastSquares}}\label{app:proof1}
For the considered $\ell_2$ objective, it holds that $\myVec{s}^{\star} = (\myMat{A}^T\myMat{A})^{-1}\myMat{A}^T\myVec{x}$. Thus, for any perturbation $\delta \in \mathbb{R}^n$,
\begin{equation}
    \|\myVec{s}_{\rm adv}^{\star}  - \myVec{s}^{\star} \|_2^2 = \|(\myMat{A}^T\myMat{A})^{-1}\myMat{A}^T\myVec{\delta} \|_2^2.
    \label{eqn:proof1}
\end{equation}
For $\myVec{\delta} \in \Delta_{2}(\epsilon)$, then \eqref{eqn:proof1} is maximized when $\myVec{\delta}$ is the scaled right singular vector of $(\myMat{A}^T\myMat{A})^{-1}\myMat{A}^T$ corresponding to its largest singular value, for which \eqref{eqn:proof1} coincides with the right hand side of \eqref{eqn:LeastSquares}. As $\Delta_2(\epsilon) \subseteq \Delta_p(\epsilon) $, $\forall p \geq 2$, the proposition holds.

\subsection{Proof of Proposition~\ref{thm:PGD}}\label{app:proof2}
We observe the $t$th iteration in \eqref{eqn:ISTA} and let $\myVec{\delta}^x$ and $\myVec{\delta}^s_t$ be the perturbations in   $\myVec{x}$ and $\myVec{s}_t$, respectively. In this case, writing $\bar{\phi}_t(\cdot) = \mu_t\rho_t\cdot \phi(\cdot)$, we obtain from \eqref{eqn:ISTA} that 
\begin{align}
     &\left\|\myVec{s}_{t+1} - {\rm Prox}_{\bar{\phi}_t}\left( \myMat{M}_t(\myVec{s}_t + \myVec{\delta}^s_t) + \mu_t\myMat{B}_t (\myVec{x}+\myVec{\delta}^x ) \right)\right\|_2\notag \\
    &=  \left\|\myVec{s}_{t+1} - {\rm Prox}_{\bar{\phi}_t}\left( \myMat{M}_t\myVec{s}_t + \mu_t\myMat{B}_t \myVec{x} + \myMat{M}_t\myVec{\delta}^s_t + \mu_t\myMat{B}_t \myVec{\delta}^x  \right) \right\|_2 \notag \\
      &\stackrel{(a)}{\leq}  \left\|\myMat{M}_t\myVec{\delta}^s_t + \mu_t\myMat{B}_t \myVec{\delta}^x  \right\|_2,
      \label{eqn:ProofBound1}
\end{align}
where $(a)$ follows from the expression of $\myVec{s}_{t+1}$ in \eqref{eqn:ISTA} combined with the non-expansiveness  property of the proximal mapping in the Euclidean space~\cite{parikh2014proximal}. 
As \eqref{eqn:ProofBound1} represents the perturbation in $\myVec{s}_{t+1}$  \eqref{eqn:ISTA}, we have that 
\begin{align}
    \|\myVec{\delta}^s_{t+1}\|_2 
    &\leq \left\|\myMat{M}_t\myVec{\delta}^s_t + \mu_t\myMat{B}_t \myVec{\delta}^x  \right\|_2 \notag \\
    &\stackrel{(b)}{\leq} \left\|\myMat{M}_t\right\|_2\cdot \left\|\myVec{\delta}^s_t\right\|_2 + \mu_t\left\|\myMat{B}_t\right\|_2\cdot \left\| \myVec{\delta}^x  \right\|_2, 
    \label{eqn:BoundRecusion}
\end{align}
where $(b)$ stems from the consistency of the $\ell_2$ induced matrix norm~\cite{bernstein2009matrix}.

We can now use the  recursion in \eqref{eqn:BoundRecusion} to characterize the difference induced at the $T$th iteration for an input perturbed by $ \myVec{\delta}^x $ with norm bounded by $\epsilon$, i.e, $ \myVec{\delta}^x \in \Delta_2(\epsilon)$. In this case, setting $\myVec{\delta}^s_0 = \myVec{0}$, we obtain by \eqref{eqn:BoundRecusion} that
\begin{equation}
   \|\myVec{\delta}^s_{T}\|_2 \leq 
   \sum_{t=0}^{T-1} \left(\prod_{j=t+1}^{T-1} \left\|\myMat{M}_j\right\|_2 \right)\mu_t\left\|\myMat{B}_t\right\|_2 \cdot \left\| \myVec{\delta}^x  \right\|_2,
\end{equation}
and thus  \ac{pgd} holds Def.~\ref{def:Lipschitz} with 
\begin{equation}
   C(\myVec{\theta}) =    \sum_{t=0}^{T-1} \left(\prod_{j=t+1}^{T-1} \left\|\myMat{M}_j\right\|_2 \right)\mu_t\left\|\myMat{B}_t\right\|_2,
\end{equation}
thus proving the proposition.

\subsection{Proof of Proposition~\ref{thm:ADMM}} \label{app:proof3}
Similarly to the previous proof, we  observe the $t$th iteration in \eqref{eqn:ADMM} and let $\myVec{\delta}^x,\myVec{\delta}^s_t,\myVec{\delta}^v_t$ and $\myVec{\delta}^y_t$ be the  perturbations in $\myVec{x},\myVec{s}_t,\myVec{v}_t$, and $\myVec{y}_t$ respectively. Using the same shorthand for the proximal parameter $\bar\phi_t=\frac{\rho_t}{2\lambda_t}\cdot\phi(\cdot)$, we can obtain:
\begin{align}
     \left\|\myVec{\delta}_{t+1}^s\right\|_2 &= \left\|\myVec{s}_{t+1} - \myMat{M}_t\left(\myVec{v}_t+\myVec{\delta}_t^v-\myVec{y}_t-\myVec{\delta}_t^y\right) -\myMat{B}_t(\myVec{x}+\myVec{\delta}^x) \right\|_2 \notag \\
     &= \left\|\myMat{M}_t(\myVec{\delta}^y_t-\myVec{\delta}^v_t)-\myMat{B}_t\myVec{\delta}^x\right\|_2 \notag \\
     &\leq  \left\|\myMat{M}_t\right\|_2\left(\left\|\myVec{\delta}^y_t\right\|_2+ \left\|\myVec{\delta}^v_t\right\|_2\right)+\left\|\myMat{B}_t\right\|_2\left\|\myVec{\delta}^x\right\|_2 . 
      \label{eqn:ProofBound_ADMM_delta_s}
\end{align}
Solving in the same manner for $\myVec{\delta}^v_t$ and $\myVec{\delta}^y_t$, we obtain:
\begin{align}
     \left\|\myVec{\delta}_{t+1}^v\right\|_2&= 
      \left\|\myVec{v}_{t+1}-{\rm Prox}_{\bar{\phi}_t}\left(\myVec{s}_{t+1}+\myVec{\delta}_{t+1}^s+\myVec{y}_t+\myVec{\delta}_t^y\right) \right\|_2\notag \\
     &\stackrel{(a)}{\leq}\left\|\delt{t+1}{s}+\delt{t}{y}\right\|_2\leq\left\|\delt{t+1}{s}
     \right\|_2+\left\|\delt{t}{y}\right\|_2,\label{eqn:ProofBound_ADMM_delta_v}\\    
     \left\|\delt{t+1}{y}\right\|_2  &= 
     \left\|\myVec{y}_{t+1}-\myVec{y}_{t}+\mu_t\left(\myVec{s}_{t+1}+\delt{t+1}{s}-\myVec{v}_{t+1}-\delt{t+1}{v} \right)\right\|_2\notag\\
     &=\left\|\mu_t\left(\delt{t+1}{s}-\delt{t+1}{v}\right)\right\|_2
     \stackrel{(b)}{\leq}|\mu_t| \left(\left\|\delt{t+1}{s}\right\|_2+\left\|\delt{t+1}{v}\right\|_2 \right)\notag\\
     &\stackrel{(c)}{\leq}|\mu_t|\left(2\normtwo{\delt{t+1}{s}}+\normtwo{\delt{t}{y}}\right),\label{eqn:ProofBound_ADMM_delta_y}
\end{align}
where $(a)$ follows from the non-expansiveness  property of the proximal mapping in the Euclidean space~\cite{parikh2014proximal}, $(b)$ stems from the consistency of the $\ell_2$ induced matrix norm~\cite{bernstein2009matrix}, and $(c)$ is derived by substituting $\normtwo{\delt{t+1}{v}}$ with the inequality in~\eqref{eqn:ProofBound_ADMM_delta_v}.
We can now use the recursive properties in \eqref{eqn:ProofBound_ADMM_delta_y} to solve for the difference induced by the auxiliary step in terms of $\delt{}{s}$ and get the following result:
\begin{align}
    \normtwo{\delt{t+1}{y}}\leq 2 \sum_{i=1}^{t+1}\normtwo{\delt{i}{s}}\prod_{j=i-1}^t|\mu_j|.
    \label{eqn:admm_dy_recursion}
\end{align}
By substituting \eqref{eqn:ProofBound_ADMM_delta_v} and \eqref{eqn:admm_dy_recursion} into \eqref{eqn:ProofBound_ADMM_delta_s} we can derive the upper-bound for $\delt{t+1}{s}$ only in terms of earlier iterations of $\delt{t}{s}$ via:
\begin{align}
  &  \normtwo{\delt{t+1}{s}} 
   \leq  \left\|\myMat{M}_t\right\|_2\left(\left\|\myVec{\delta}^y_t\right\|_2+ \left\|\myVec{\delta}^s_t\right\|_2 \!+\! \left\|\myVec{\delta}^y_{t-1}\right\|_2\right)\!+\!\left\|\myMat{B}_t\right\|_2\left\|\myVec{\delta}^x\right\|_2 \notag \\
     &\leq  \left\|\myMat{M}_t\right\|_2\left(2|\mu_{t-1}|\left\|\myVec{\delta}^s_t\right\|_2\!+\! |\mu_{t-1}|\left\|\myVec{\delta}^y_{t-1}\right\|_2 \!+ \!\left\|\myVec{\delta}^s_t\right\|_2 \!+ \!\left\|\myVec{\delta}^y_{t-1}\right\|_2\right) \notag \\
     &\qquad +\left\|\myMat{B}_t\right\|_2\left\|\myVec{\delta}^x\right\|_2 \notag \\ 
     &\leq  \normtwo{\myMat{M}_t}\Big(\notag
     \left(2|\mu_{t-1}|\!+\!1\right)\normtwo{\delt{t}{s}}\!+\!
     \left(|\mu_{t-1}|\!+\!1\right)\normtwo{\delt{t-1}{y}}
     \Big)\notag \\
     &\qquad + \normtwo{\myMat{B}_t}\normtwo{\delt{}{x}}\notag \\
     &\leq  \normtwo{\myMat{M}_t}\Big(\notag
     2\left(|\mu_{t-1}|\!+\!1\right)\normtwo{\delt{t}{s}}\!+\!
     2\left(|\mu_{t-1}|\!+\!1\right)\notag \\
     &\qquad \times \sum_{i=1}^{t-1}\normtwo{\delt{i}{s}}\prod_{j=i-1}^{t-2}|\mu_j|
     \Big) \! -\! \normtwo{\myMat{M}_t}\normtwo{\delt{t}{s}} \!+\! \normtwo{\myMat{B}_t}\normtwo{\delt{}{x}}\notag \\ 
    &\leq
   \normtwo{\myMat{M}_t}\cdot 2\left(|\mu_{{t-1}}|+1\right)\Big(\normtwo{\delt{t}{s}}+\sum_{i=1}^{t-1}\normtwo{\delt{i}{s}}\prod_{j=i-1}^{t-2}|\mu_j|\Big)
    \notag \\
     &\qquad +\normtwo{\myMat{B}_t}\cdot\normtwo{\delt{t}{x}}
    =\normtwo{\myMat{B}_t}\cdot\normtwo{\delt{}{x}}+\sum_{i=1}^t\myVec{w}_i^t\normtwo{\delt{i}{s}} 
     \notag\\
     &
    \stackrel{(d)}{\leq}\normtwo{\myMat{B}_t^*}\cdot\normtwo{\delt{}{x}} \prod_{i=0}^t(1+\myVec{w}_i^t),
    \label{eqn:admm_proof_s_recursion}
\end{align} 
where we define $\myVec{w}^t_i=2\normtwo{\myMat{M}_t}\cdot\left(|\mu_{t-1}|+1\right)\prod_{j=i-1}^{t-2}|\mu_j|$ in which the product would be equal to 1 if the index set is empty. Also $(d)$ stems from the discrete Gronwall's Inequality  ~\cite{clark1987short}, using $\normtwo{\myMat{B}_t^*}=\max_{0 \leq i \leq t}{\normtwo{\myMat{B}_i}}$.

\bibliographystyle{IEEEtran}
\bibliography{example_paper}

\end{document}